\documentclass[sigconf]{acmart}
\usepackage{algorithm}
\usepackage{color}
\usepackage{listings}
\usepackage{array}
\usepackage{bbding}
\newcommand{\ts}{\textsuperscript}

\AtBeginDocument{%
  \providecommand\BibTeX{{%
    \normalfont B\kern-0.5em{\scshape i\kern-0.25em b}\kern-0.8em\TeX}}}

\copyrightyear{2022}
\acmYear{2022}
\setcopyright{acmcopyright}\acmConference[MM '22]{Proceedings of the 30th ACM
International Conference on Multimedia}{October 10--14, 2022}{Lisboa, Portugal}
\acmBooktitle{Proceedings of the 30th ACM International Conference on Multimedia
(MM '22), October 10--14, 2022, Lisboa, Portugal}
\acmPrice{15.00}
\acmDOI{10.1145/3503161.3547934}
\acmISBN{978-1-4503-9203-7/22/10}

\acmSubmissionID{764}

\begin{document}

\title{Real-time Streaming Video Denoising with Bidirectional Buffers}

\author{Chenyang Qi}
\authornotemark[1]
\affiliation{%
  \institution{HKUST}
}

\author{Junming Chen}
\authornote{Both authors contributed equally to this research.}
\affiliation{%
  \institution{HKUST}
}

\author{Xin Yang}
\affiliation{%
  \institution{HKUST}
}

\author{Qifeng Chen}
\affiliation{%
  \institution{HKUST}
}


\begin{abstract}


Video streams are delivered continuously to save the cost of storage and device memory. 
Real-time denoising algorithms are typically adopted on the user device to remove the noise involved during the shooting and transmission of video streams.
However, sliding-window-based methods feed multiple input frames for a single output and lack computation efficiency. Recent multi-output inference works propagate the bidirectional temporal feature with a parallel or recurrent framework, which either suffers from performance drops on the temporal edges of clips or can not achieve online inference.
In this paper, we propose a Bidirectional Streaming Video Denoising (BSVD) framework, to achieve high-fidelity real-time denoising for streaming videos with both past and future temporal receptive fields.
The bidirectional temporal fusion for online inference is considered not applicable in the MoViNet. However, we introduce a novel Bidirectional Buffer Block as the core module of our BSVD, which makes it possible during our pipeline-style inference.
In addition, our method is concise and flexible to be utilized in both non-blind and blind video denoising. We compare our model with various state-of-the-art video denoising models qualitatively and quantitatively on synthetic and real noise. Our method outperforms previous methods in terms of restoration fidelity and runtime. Our source code is publicly available at \url{https://github.com/ChenyangQiQi/BSVD}

\end{abstract}



\maketitle

\vspace{-5pt}
\section{Introduction}

With the explosive growth of social media, there is an increasing need for video streaming applications on the user's device, such as live streaming on YouTube, TikTok, and video meetings on Zoom. 
Since the noise from video capturing and compression degrades the quality of videos, noise reduction plays a vital role in improving video fidelity. 
Although offline video processing~\cite{Vaksman2021Patch,chan2021basicvsr,liang2022vrt,tassano2019dvdnet} has been an active research topic in recent years, few works~\cite{Tassano2020FastDVDNet,Maggioni2021Efficient,Wen2017Joint,Wen2019VIDOSAT} are feasible for online streaming video denoising, which is more challenging.
Streaming video denoising is featured in the following two points compared with offline video denoising.
First, since streaming video denoising is primarily applied to sports games and other live streaming media, a real-time inference speed should be satisfied.
Second, the frames should be processed online in a continuous frame-by-frame way because there is no explicit end for streaming video.
To fulfill the above requirements, we propose a real-time high-quality streaming denoising framework.

\begin{figure}[t]
\centering
\includegraphics[width=\linewidth]{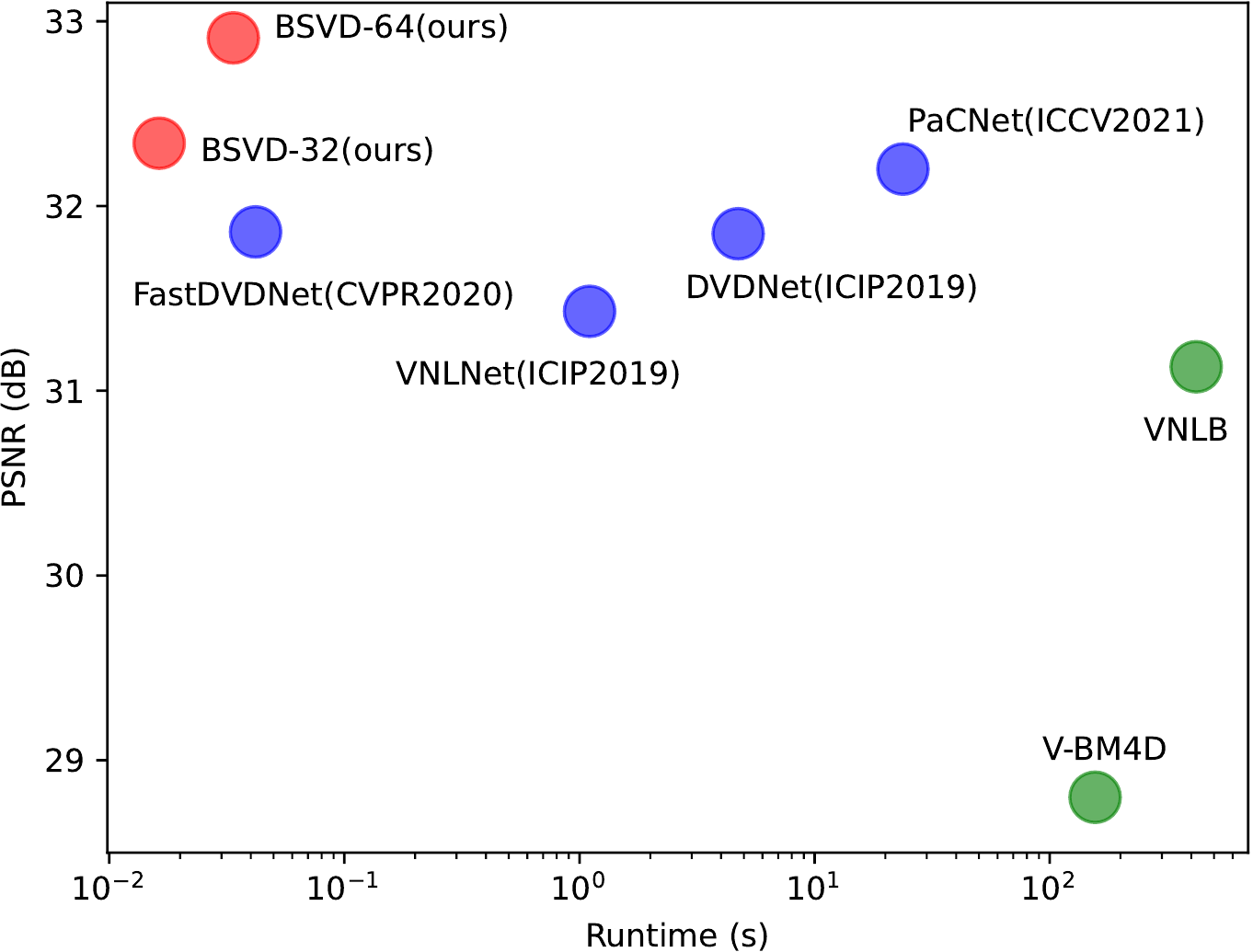}
\caption{\textbf{Comparison of PSNR and runtime on DAVIS test set with noise level $\sigma=50$}. Our BSVD-64 outperforms state-of-the-art PaCNet~\cite{Vaksman2021Patch} with $700\times$ speedup. 
It takes BSVD-64 33.7ms per frame to process a video with resolution of $960\times540$.
\textcolor{teal}{Green}: CPU method. \textcolor{blue}{Blue}: Learning-based GPU method. \textcolor{red}{Red}: Ours.}
\label{fig:teaser}
\vspace{-1em}
\end{figure}

\begin{figure*}[t]
\centering
\resizebox{\textwidth}{!}{
\begin{tabular}{@{}m{3cm}<{\raggedright}*{4}{m{3cm}<{\centering}}m{4.5cm}<{\centering} }
\toprule
Method & (a) Sliding-window & (b) Uni-RNN & (c) Bi-RNN & (d) MIMO & (e) Ours \\
\midrule
\includegraphics[width=1.0\linewidth]{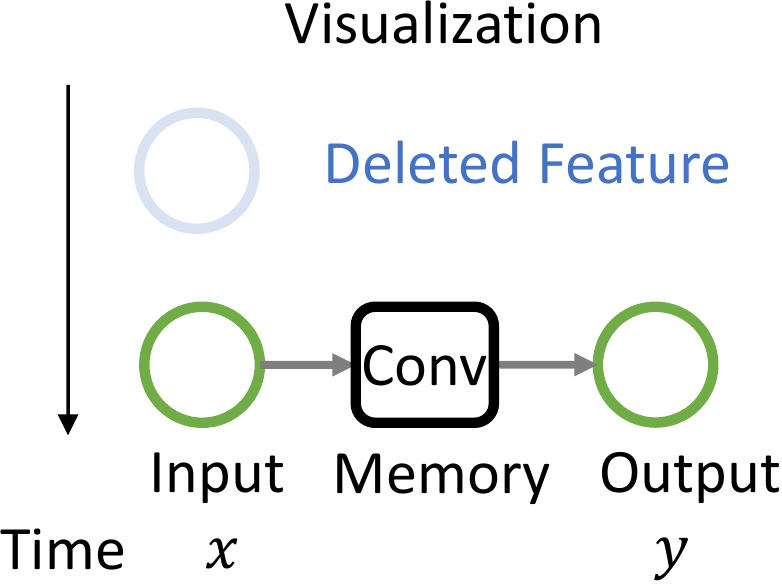}&
\includegraphics[width=\linewidth]{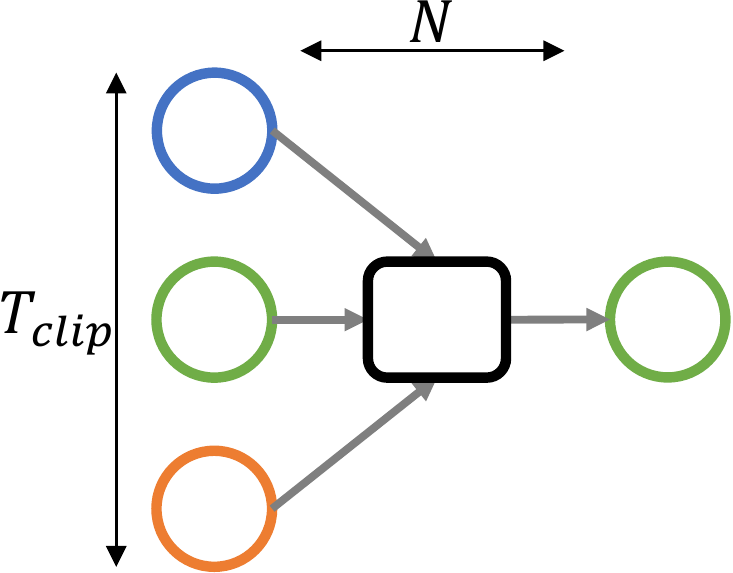}&
\includegraphics[width=0.8\linewidth]{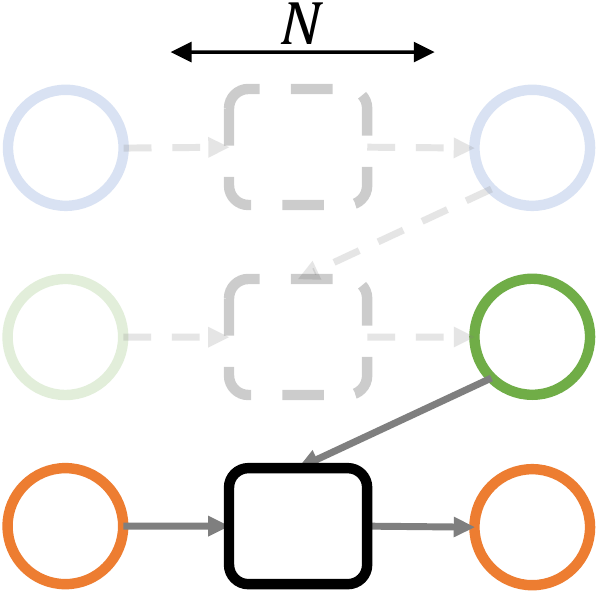}&
\includegraphics[width=\linewidth]{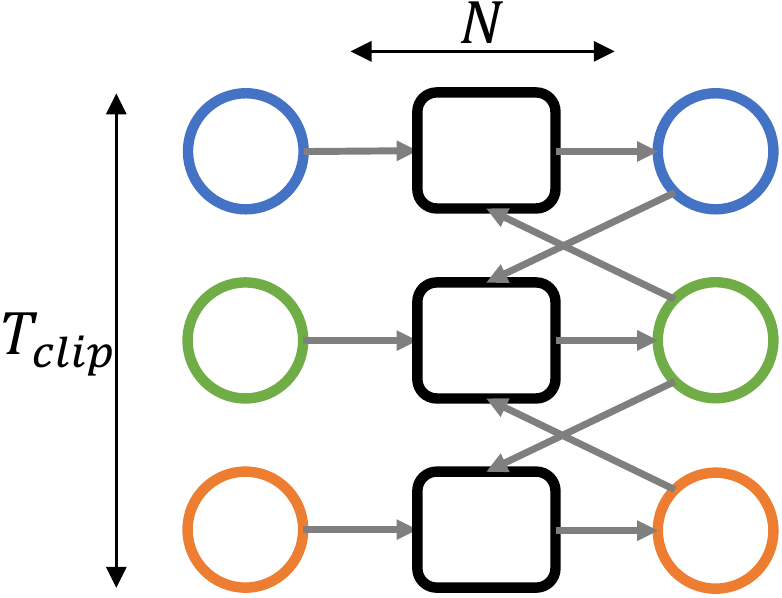}&
\includegraphics[width=\linewidth]{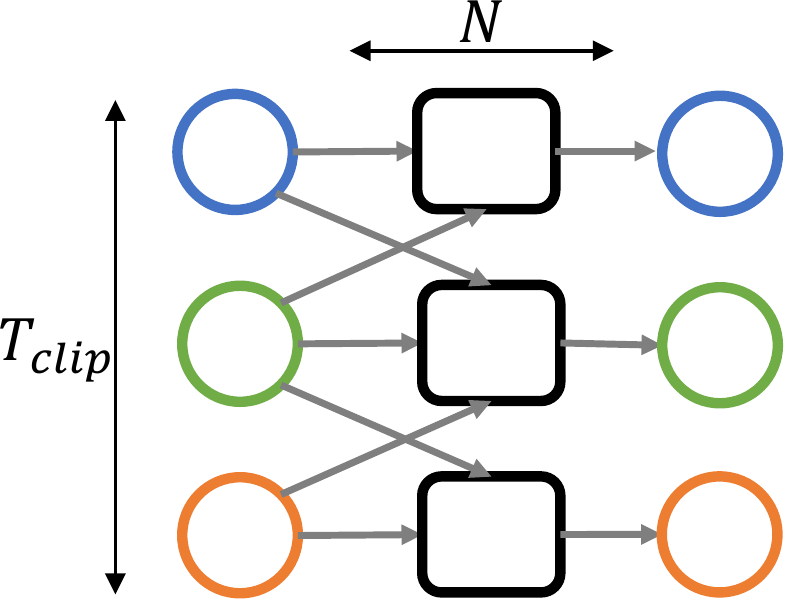}&
\includegraphics[width=0.9\linewidth]{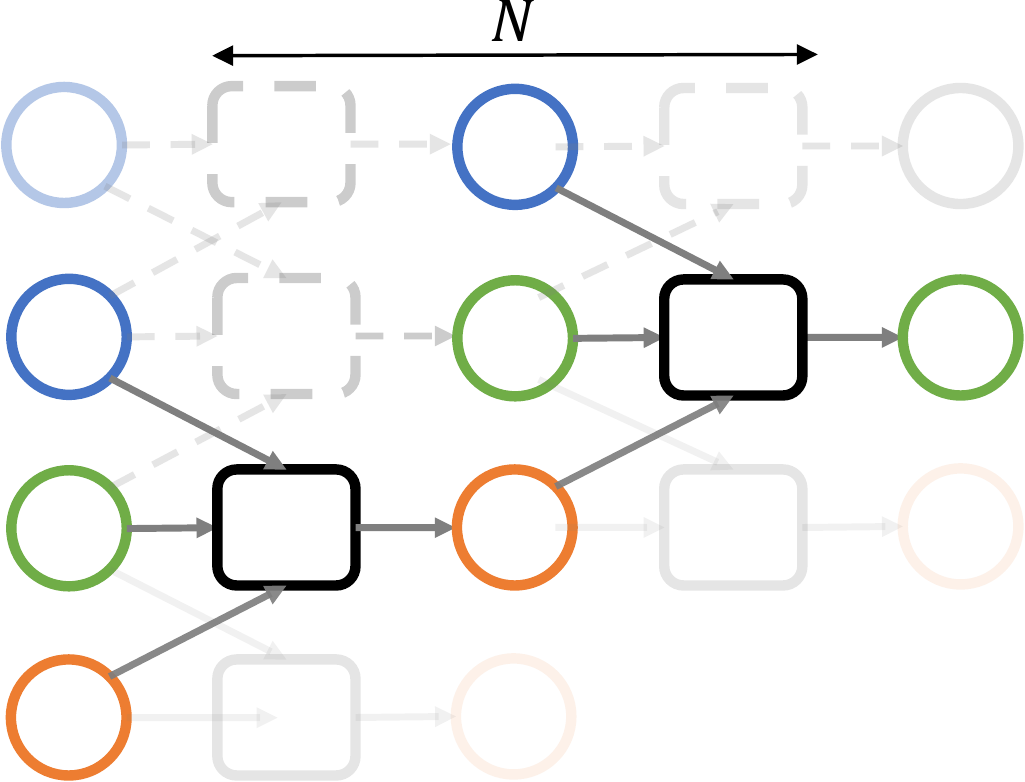}\\
\midrule
{Input} & {$x_{i-T_{clip}/2}, \ldots, x_{i+T_{clip}/2}$} & {$x_{i}$} &  {$x_{1},x_{2},\hdots,x_{T_{clip}}$} & {$x_{1},x_{2},\hdots,x_{T_{clip}}$} & {$x_{i}$} \\
{Buffered Feature Index} & {-} & {$i-1$} & {$1,2,\hdots,T_{clip}$} & {-} & {$i-N,\hdots,i-1$} \\
{Output} & {$y_{i}$} &  {$y_{i}$} & {$y_{1}, y_{2}, \hdots y_{T_{clip}}$} & {$y_{1}, y_{2}, \hdots y_{T_{clip}}$} & {$y_{i-N}$} \\
\midrule
{Memory Complexity} & {$O(T_{clip})$} & $O(N)$ & {$O(T_{clip})$} & {$O(T_{clip})$}& {$O(N)$} \\
\specialrule{0em}{1pt}{1pt}
{Time Per Frame} 
& {$O(T_{clip}N)$} & $O(N)$ & $O(N)$ & $O(N)$ & $O(N)$ \\
\specialrule{0em}{1pt}{1pt}

{Bidirectional Fusion}  & \textcolor{teal}{\Checkmark} &\textcolor{red}{\XSolidBrush}    & \textcolor{teal}{\Checkmark} & \textcolor{teal}{\Checkmark} & \textcolor{teal}{\Checkmark}\\
{Clip-edge Fidelity} & \textcolor{teal}{\Checkmark}& \textcolor{teal}{\Checkmark}&\textcolor{red}{\XSolidBrush}& \textcolor{red}{\XSolidBrush} & \textcolor{teal}{\Checkmark}\\


\bottomrule
\end{tabular}
}

\caption{\textbf{The comparison of computation graph and complexity for different methods.} For an input clip with length $T_{clip}$, we assume all methods use same $N$ convolution blocks for temporal fusion. 
\textcolor{blue}{Blue}, \textcolor{teal}{green}, and \textcolor{red}{red} features represent the past, present, and future features from three adjacent frames.
Solid features are cached in the GPU memory, while dotted features have been deleted.
Compared with sliding-window methods (a), our inference time is shorter. For unidirectional-RNN (b), our framework utilize bidirectional temporal fusion, and achieve better fidelity. In addition, bidirectional-RNN (c) and MIMO (d) framework suffer from $O(T_{clip})$ memory and fidelity degradation on the clip edges, which is solved in our inference framework.
}
\vspace{-8pt}
\label{fig:comp_graph}
\end{figure*}

To the best of our knowledge, the existing works can be categorized into three classes: sliding-window-based methods~\cite{Tassano2020FastDVDNet,Vaksman2021Patch} (Fig.~\ref{fig:comp_graph}(a)), 
recurrent methods~\cite{Maggioni2021Efficient,Xiang2022ReMoNet} (Fig.~\ref{fig:comp_graph}(b,c)) and multi-input multi-output (MIMO) methods~\cite{Xiang2022ReMoNet,liang2022vrt} (Fig.~\ref{fig:comp_graph}(d)).
The sliding-window-based methods~\cite{Tassano2020FastDVDNet,Vaksman2021Patch} (Figure~\ref{fig:comp_graph}(a)) restore each frame by feeding the degraded one with its neighbours. 
However, such sliding-window strategy has unnecessary, redundant computation: each frame is fed into the network multiple times. 
For instance, to restore a video with resolution of $960 \times 540$, it takes PaCNet~\cite{Vaksman2021Patch} about 24 seconds per frame, which impedes this method to real-time application. 
Meanwhile, such methods only fuse the temporal information in pixel-level~\cite{Tassano2020FastDVDNet} or neighbor patch-level~\cite{Vaksman2021Patch}, while the intermediate multi-scale features are also important for the denoising performance, as shown in (Sec.~\ref{section:Bidirectional_feature_propagation}).

Different from sliding-window-based methods, recurrent works  \cite{Maggioni2021Efficient,chan2021basicvsr} use previous reconstructed results as a reference for next restoration. Such recurrent methods can be further divided into unidirectional~\cite{Maggioni2021Efficient,Xiang2022ReMoNet} (Fig.~\ref{fig:comp_graph}(b)) and bidirectional~\cite{chan2021basicvsr} (Fig.~\ref{fig:comp_graph}(c)) propagation. Although unidirectional recurrent methods can exploit past information for streaming processing, their denoising performance degrades by ignoring future information. To utilize both past and future information, bidirectional recurrent methods are proposed to propagate the temporal information in both forward and backward directions. For best results from a bidirectional recurrent network, users must prepare the entire video in advance, which means they cannot realize streaming video denoising~\cite{chan2021basicvsr}.

Most recently, MIMO frameworks~\cite{Xiang2022ReMoNet,liang2022vrt} have been introduced for efficient processing a clip of video frames
in one-forward-pass(Fig.~\ref{fig:comp_graph}(d)). The inference schema can be implemented as a channel shift~\cite{Lin2019TSM} or a local window~\cite{liang2022vrt} in temporal dimension. However, with the growth of clip size $T_{clip}$, the inference-time memory consumption also increases linearly in $\mathcal{O}\left(T_{clip}\right)$.
Thus, such methods suffer from heavy memory consumption~\cite{liang2022vrt}, and the long input video must be segmented into short clips.
Moreover, similar to bidirectional recurrent methods~\cite{chan2021basicvsr}, MIMO also has
performance degradation on the edges of clips (Sec.~\ref{section:Streaming inference vs parallel inference}).

In this work, we propose the \textbf{B}idirectional \textbf{S}treaming \textbf{V}ideo \textbf{D}enoising framework (BSVD). 
Bidirectional temporal fusion is critical for low-level video processing since it fully utilizes the information of both past and future frames, which is demonstrated in our experiment (Sec.~\ref{section:Bidirectional_feature_propagation}). Thus, we train our model as a MIMO framework using bidirectional Temporal Shift Modules (TSM)~\cite{Lin2019TSM}, which is also more efficient than sliding-window.
To address the clip-edges-drop problem~\cite{chan2021basicvsr,liang2022vrt} and enable the bidirectional fusion for streaming videos, which is regarded as not applicable in previous works~\cite{Lin2019TSM,Kondratyuk2021movinets}, we propose a Bidirectional Buffer Block that can cache and reuse the features for both past and future propagation during pipeline-style inference.
Thus, our BSVD featured with the pipeline-style buffer block can achieve a constant memory complexity of $\mathcal{O}\left(1\right)$, in contrast with $\mathcal{O}\left(T_{clip}\right)$ in other MIMO methods~\cite{chan2021basicvsr, liang2022vrt}. Extensive evaluations on datasets with synthetic and real noise show that the proposed method outperforms previous methods in denoising performance and runtime.


Our contributions can be summarized as follows:

\textbf{(a)} We propose a pipeline-style buffer-based video denoising framework, named BSVD, which processes the video streams of resolution $960\times540$ in real-time.

\textbf{(b)} 
The proposed novel Bidirectional Buffer Block and pipeline-style inference framework enable bidirectional temporal fusion for online streaming video processing.

\textbf{(c)}
In addition, we address the fidelity degradation on the clip edges in the MIMO framework~\cite{chan2021basicvsr} by
utilizing buffered features.
Our proposed computation graph is general and applicable for pre-trained checkpoints of the existing method FastDVDNet~\cite{Tassano2020FastDVDNet}.
\begin{figure*}[t]
\centering
\resizebox{\textwidth}{!}{
\begin{tabular}{m{6.5cm}<{\centering} m{3.5cm}<{\centering} m{6.5cm}<{\centering}}
\includegraphics[width=\linewidth]{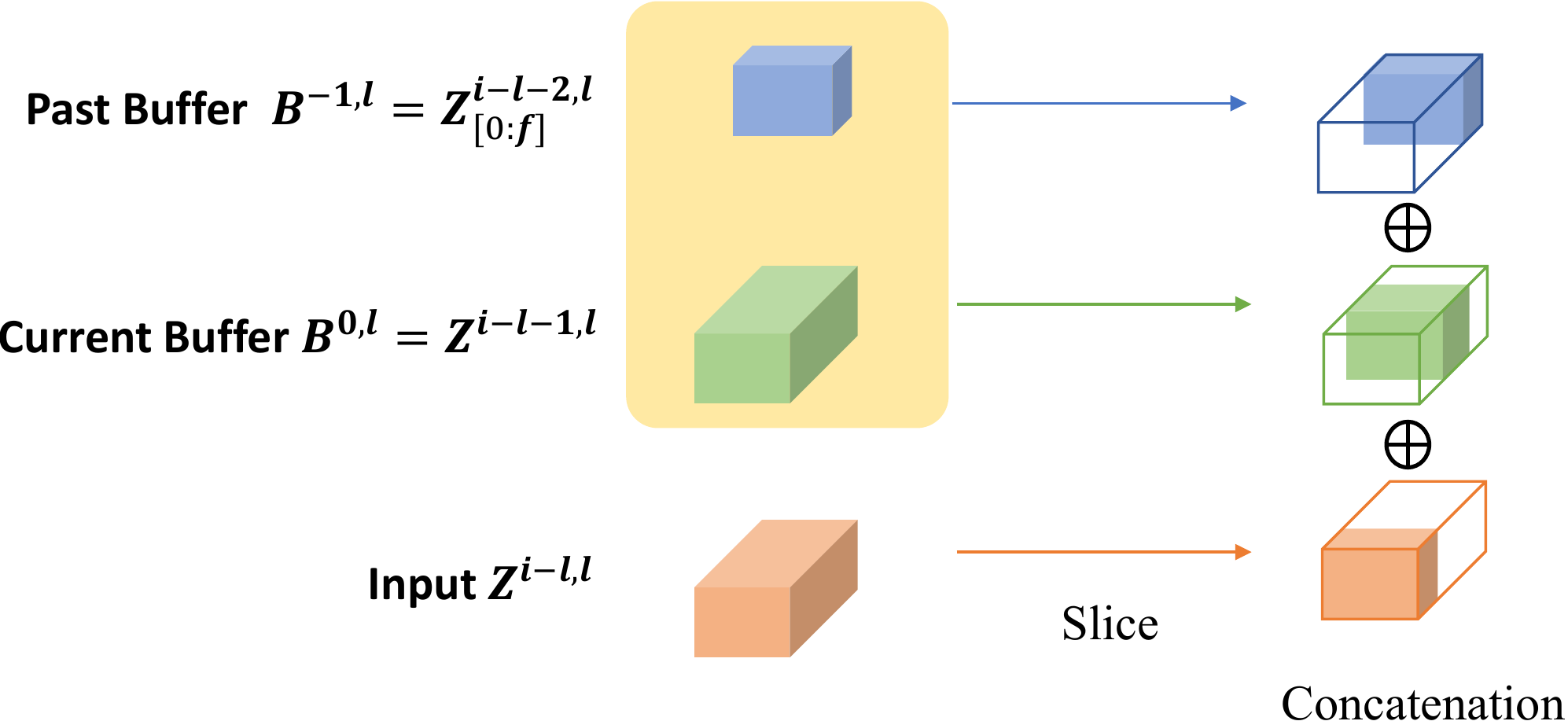}&
\includegraphics[width=\linewidth]{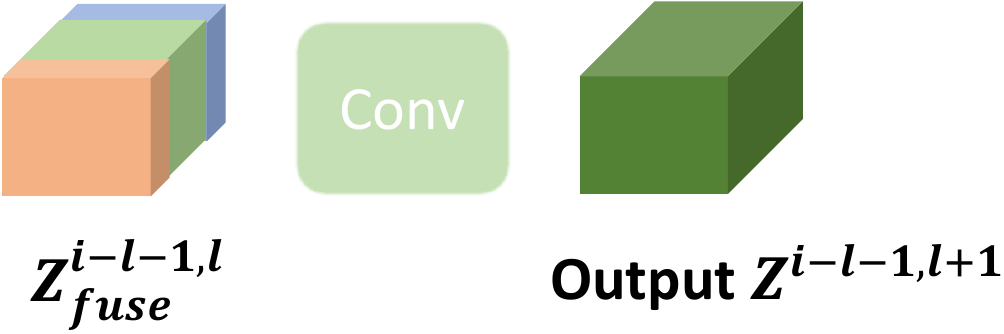}&
\includegraphics[width=\linewidth]{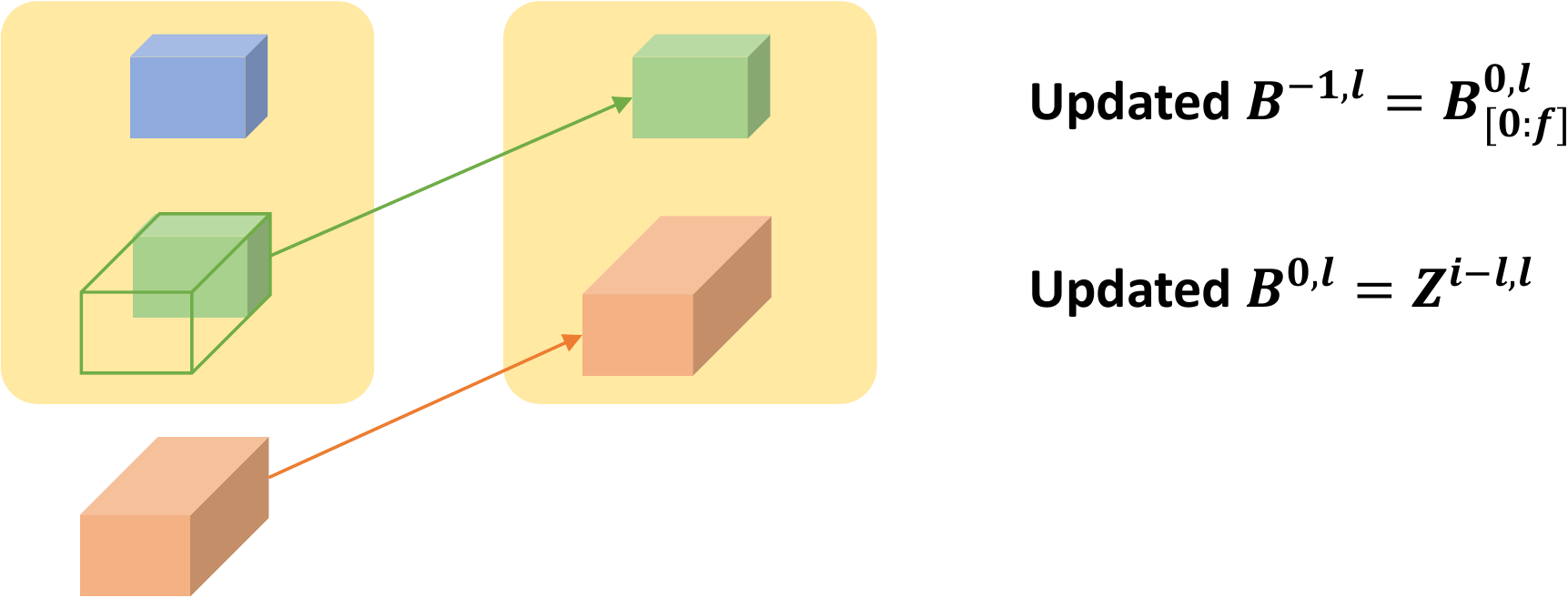}
\\
(a) Temporal Fusion & (b) 2D Convolution & (c) Update Buffer \\
\end{tabular}
}

\caption{The forward operation of $l$\ts{th} Bidirectional Buffer Block at time step $i$. This block aggregates input feature $Z^{i-l,l}$ with buffered features $B^{-1,l}, B^{0,l}$, to output $Z^{i-l-1, l+1}$, which is the input for the $(l+1)$\ts{th} temporal buffer block. After convolution operation, the buffered feature are updated using input $Z^{i-l,l}$.}

\label{fig:Bidirectional Buffer Block}
\end{figure*}
\begin{figure*}[t]
\centering
\includegraphics[width=0.9\linewidth]{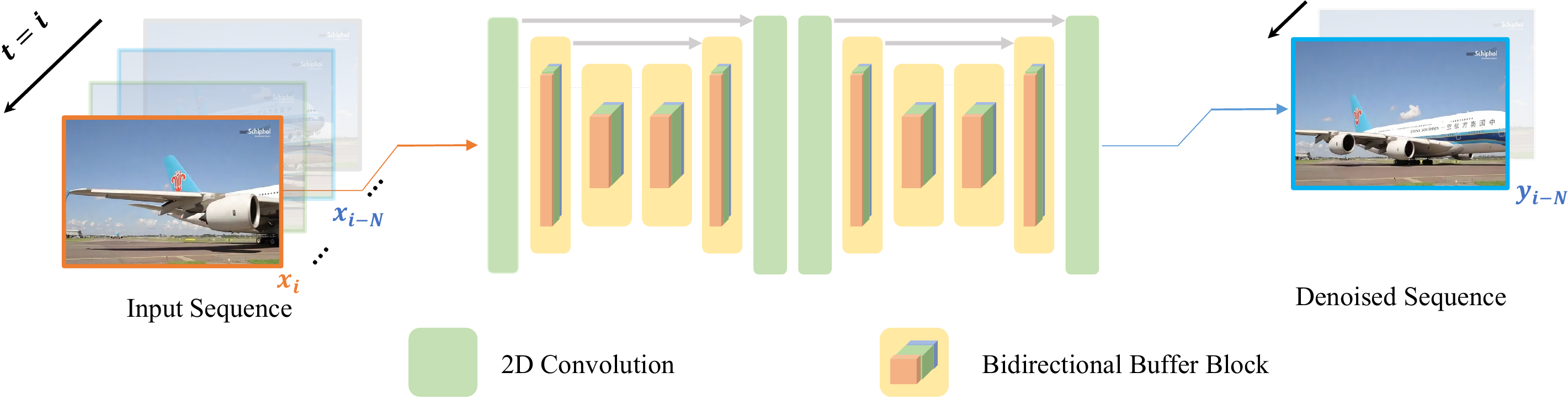}
\caption{ \textbf{An overview of our framework}. The backbone of our network is two light-weight U-Nets~\cite{Ronneberger2015U-Net} with temporal fusion operation inserted between convolution layers. At time step $i$ during inference, one noisy frame $x_i$ and its noise map are fed into the neural network. Then, our network outputs another clean frame $y_{i-N}$.}
\label{fig:overview}
\end{figure*}

\textbf{(d)} Our succinct BSVD framework 
achieves SOTA performance on Gaussian noise~\cite{Tassano2020FastDVDNet,Vaksman2021Patch}, real noise~\cite{Maggioni2021Efficient}, and blind denoising~\cite{Sheth2021UDVD,Xiang2022ReMoNet}. Figure~\ref{fig:teaser} provides a comparison of PSNR and runtime on DAVIS test set with noise level $\sigma=50$. BSVD outperforms the previous most efficient baseline FastDVDnet~\cite{Tassano2020FastDVDNet} 
with more than $2\times$ speedup and improves average PSNR by 0.57dB on DAVIS dataset. Moreover, it surpasses the best fidelity work PaCNet~\cite{Vaksman2021Patch} while being 700$\times$ faster.

\section{Related Work}

\textbf{Image denoising. }
Single image denoising has been a long-standing problem as one of the most fundamental image restoration tasks. Some traditional and representative methods~\cite{Dabov2007imagedenoising,Maggioni2013NonlocalTF,Buades2011Non-Local,Lebrun2013Nonlocal} typically exploit the texture similarity in non-local image patches to achieve satisfying denoising performance. However, such methods suffer from time-consuming patch searching. Recently, deep learning has made great progress for single image denoising tasks. Zhang et al.~\cite{zhang2017beyond} propose a deep CNN using residual learning, which outperforms the traditional methods by a large margin. They present an efficient model using downsampled sub-images and a non-uniform noise level map. Meanwhile, some other works attempt to predict pixel-wise kernels~\cite{Mildenhall2018Burst,Xia2020Basis,Lu2021Efficient} firstly and then apply these spatial-variant kernels as convolutional weights in the main denoising network branch. 
Recently, Transformer-based architectures~\cite{liang2021swinir,Zamir2021Restormer} achieve state-of-the-art performance on single image denoising tasks.
While single-image methods can restore each frame online, they also discard the information in adjacent frames, which should be considered in the streaming denoising task.

\textbf{Video denoising.}
In the video, pixels in adjacent frames can be very similar. The temporal correlation can be utilized by deformable convolution~\cite{Wang2019EDVR}, kernel prediction~\cite{Mildenhall2018Burst,Xia2020Basis}, block matching~\cite{Maggioni2012VBM4D,Vaksman2021Patch}, optical flow~\cite{tassano2019dvdnet} and attention~\cite{liang2022vrt} which typically improve the image fidelity 
at the cost of expensive computation.
Meanwhile, various works focus on the reduction of computation in video denoising~\cite{Tassano2020FastDVDNet,Maggioni2021Efficient}. FastDVDnet~\cite{Tassano2020FastDVDNet} is composed of two lightweight U-Nets~\cite{Ronneberger2015U-Net} without explicit motion estimation. 
It works in a multiple-to-single manner, while ours adopt an efficient buffer-based inference pipeline without redundancy. 
More recently, EMVD~\cite{Maggioni2021Efficient} presents a unidirectional recurrent solution, which reuses the previous predicted clean frame.
Compared with EMVD~\cite{Maggioni2021Efficient}, our buffer-based bidirectional fusion fully utilized the feature from neighboring frames. 
Meanwhile, we solve the problem of high memory consumption and performance drop on the temporal edges of clips, which widely exist in MIMO frameworks \cite{Xiang2022ReMoNet, liang2022vrt, chan2021basicvsr}.


\section{Method}

Given a stream of noisy frames without an explicit ending frame $\{x_0, x_1, \hdots,x_i,\hdots\}$, where $x_i \in \mathbb{R}^{ C\times H \times W }$ with input channel $C$, height $H$ and width $W$, we restore the clean frames $\{y_0, y_1, \hdots,y_i,\hdots\}$ in a continuous pipeline. 
Our method is suitable for both non-blind and blind denoising. 
For non-blind denoising, we set $C=4$~\cite{Vaksman2021Patch,Tassano2020FastDVDNet} or $C=5$ ~\cite{Maggioni2021Efficient} in input frames following previous works. For blind denoising, we only input the RGB channels and thus $C=3$.
We train our model in MIMO framework using Temporal Shift Module(TSM)~\cite{Lin2019TSM} for distributed parallelism (Sec.~\ref{section:training_tsm}). During inference, 
we replace TSM with Bidirectional Buffer Block
(Sec.~\ref{section:Bidirectional Buffer Block}), and our framework processes the video in a pipeline (Sec.~\ref{section:Pipeline-style Inference Scheme}).
Figure~\ref{fig:comp_graph} 
demonstrates the difference between our method and previous inference frameworks.
Figure~\ref{fig:overview} gives an overview of our inference algorithm with Bidirectional Buffer Block (Figure~\ref{fig:Bidirectional Buffer Block}).
 Specifically, at time step $i\in\{0,1,2,\hdots\}$, we feed a noisy frame $x_i$. Then, the input is fused with previous features buffered in $N$ Bidirectional Buffer Blocks. Finally, the model produces a single clean frame {$y_{i-N}$} for the noisy input at time step $i-N$ at the end of the time step $i$.

\subsection{Training with Temporal Shift Module}
\label{section:training_tsm}
\begin{figure}[t]
\center
\includegraphics[width=0.8\linewidth]{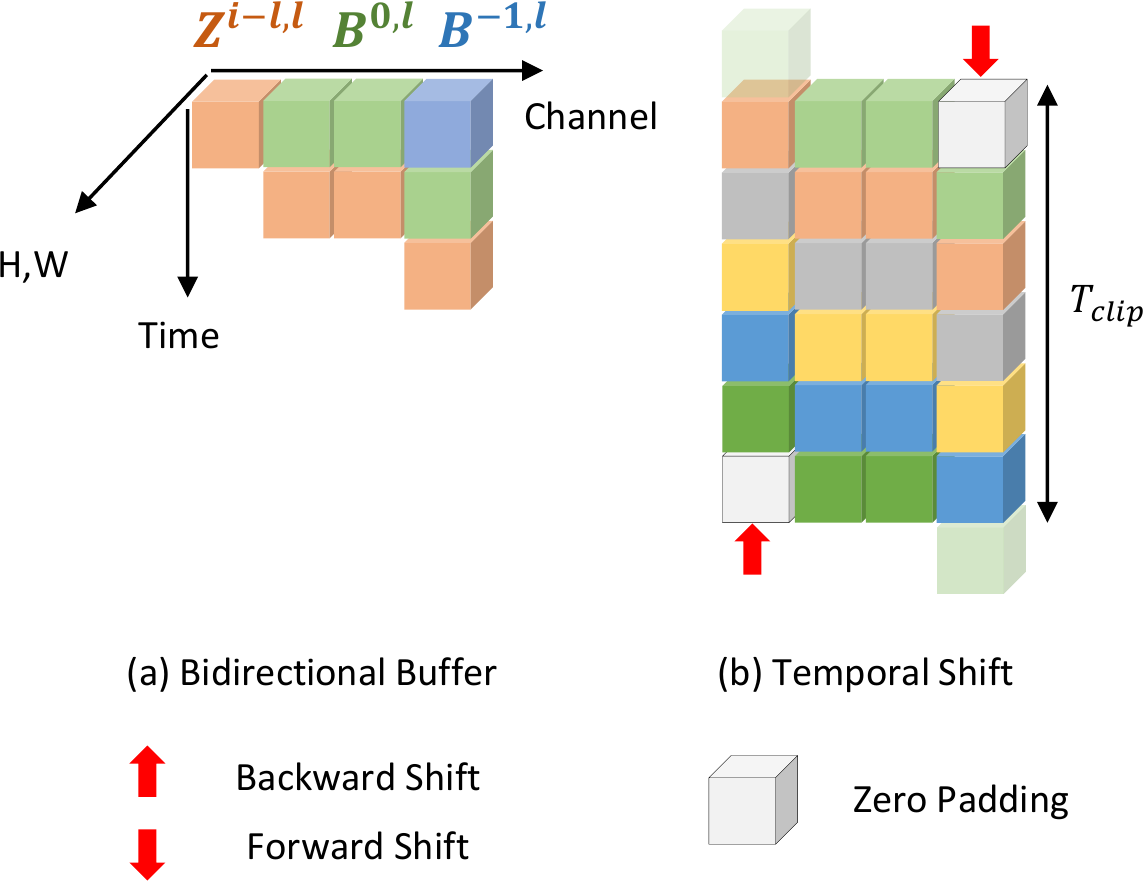}
\caption{Illustration of the difference between a single block of our pipeline-style inference method and TSM. We set the ratio $r=4$ in visualization. TSM uses zero paddings to fill the slots produced by shift operation, while we use buffered feature without zero padding at the temporal boundaries. \vspace{-15pt}}
\label{fig:temporal_shift}
\end{figure}



During the training stage, we utilize TSM~\cite{Lin2019TSM}
as the temporal fusion operation in our convolution backbone.
We randomly crop $T_{clip}$-frame clips from videos as input and use the pixel loss (e.g., L1 or L2) between 
$\{x_t\}_{t=1}^{T_{clip}}$ 
and 
$\{y_t\}_{t=1}^{T_{clip}}$
as the training objective. Assume a feature clip $Z \in \mathbb{R}^ {T_{clip}  \times C_f \times H_f \times W_f}$ is the stack of $T_{clip}$ continuous frame features, with feature channel $C_f$, height $H_f$ and width $W_f$.
Then, the temporal  shift can be represented as the concatenation ($\oplus$) of features channels:
\begin{eqnarray}
Z^0_{fused} =&  Z _ {[0:f]} ^ {-1} \oplus Z _ {[f:-f]} ^ {0} \oplus Z _ {[-f:]} ^ {+1},
\label{eqn:tsm}
\end{eqnarray}
where
superscript $-1$ and $+1$ represent one-frame forward and backward shift, respectively. $f = \lfloor C_f/r\rfloor$ is the number of shifted channels, and the ratio of channels shifted in each direction is empirically set as $r=8$ following TSM~\cite{Lin2019TSM}.
As shown in Figure~\ref{fig:temporal_shift}(b), TSM is a MIMO framework, which generates blank slots and humps on the boundaries of the clip.
The original TSM~\cite{Lin2019TSM} directly truncates the humps and fills the blank slots with zero paddings, which degrades the image fidelity. Thus, it is not an optimal inference framework for processing video streams.

\textbf{Backbone network.}
We utilize two enhanced lightweight U-Nets~\cite{Xia2017WNetAD,Ronneberger2015U-Net} as our base model W-Net~\cite{Xia2017WNetAD}.
Such a two-step denoising architecture has proven to be effective in previous works~\cite{Tassano2020FastDVDNet, Xia2017WNetAD,Sheth2021UDVD}. Batch normalization (BN) layer is observed to decrease image fidelity in super-resolution~\cite{Lim2017EDSR,esrgan} and deblurring~\cite{SeungjunNah2017DeepMC}, but it still exists in current SOTA video denoising methods~\cite{Tassano2020FastDVDNet,Vaksman2021Patch}.
Thus, we remove BN layers in the W-Net.
Besides, we replace ReLU activation with ReLU6 to alleviate artifacts during FP16 inference. In the following, ``BSVD-32'' denotes our model whose smallest feature channel is ``32''. We adjust our backbone by multiplying or dividing all channel numbers by a scale. Please refer to the supplement for more details about our implementation.


\begin{algorithm}[t]
\caption{\small{Pseudo code of a single Bidirectional Buffer Block.}}
\label{alg:pseudocode}
\definecolor{codeblue}{rgb}{0.25,0.5,0.5}
\definecolor{codered}{rgb}{0.75,0.25,0.25}
\lstset{
	backgroundcolor=\color{white},
	basicstyle=\fontsize{7.2pt}{7.2pt}\ttfamily\selectfont,
	columns=fullflexible,
	breaklines=true,
	captionpos=b,
	commentstyle=\fontsize{7.2pt}{7.2pt}\color{codeblue},
	keywordstyle=\fontsize{7.2pt}{7.2pt}\color{codered},,
}
\vskip -0.075in
\begin{lstlisting}[language=python,mathescape=true]
class BidirectionBufferedBlock(nn.Module):
    def __init__(self, in_channels, out_channels, r=8):
        self.conv = nn.Conv2d(in_channels, out_channels)
        self.f = in_channels//r
        self.$B^{-1,l}$ = torch.zeros(n, self.f, h, w)
        self.$B^{0,l}$  = torch.zeros(n, in_channels, h, w)
    def forward(self, $Z^{i-l,l}$):
        # Activate the block
        if self.$B^{0,l}$ is None: 
            self.$B^{0,l}$ = $Z^{i-l,l}$
            return None
        # Fuse the temporal adjacent features
        fusion = torch.cat([
            self.$B^{-1,l}$,
            self.$B^{0,l}$[:,self.f:-self.f, :, :]
            $Z^{i-l,l}$[:, -self.f:, :, :], dim=1)
        # 2D Convolution for output
        $Z^{i-l-1,l+1}$ = self.conv(fusion)
        # Update the buffer
        self.$B^{-1,l}$ = self.$B^{0,l}$[:, :self.f, :, :]
        self.$B^{0,l}$ = $Z^{i-l,l}$
        return $Z^{i-l-1,l+1}$
\end{lstlisting}
\vspace{-6pt}
\end{algorithm}

\begin{table}[t]
\vspace{-25pt}
\end{table}

\subsection{Inference with Bidirectional Buffer Block }
\label{section:Bidirectional Buffer Block}
\label{section:Bidirectional Propagation Block}

During the inference stage, we replace the temporal shift operation with the designed Bidirectional Buffer Block, while keeping the weights of the convolution backbone unchanged.
As shown in Figure~\ref{fig:temporal_shift}(a), our buffer blocks serve as a bridge to transfer the feature of neighboring frames, which can enlarge the temporal receptive field as well as make bidirectional temporal fusion possible for streaming video, which is regarded not applicable in previous works~\cite{Lin2019TSM,Kondratyuk2021movinets}. Therefore, we solve the fidelity degradation at two clip boundaries in such a MIMO system.

The pseudo code shown in Alg.~\ref{alg:pseudocode} and Figure~\ref{fig:Bidirectional Buffer Block} delineate the computation steps of a single Bidirectional Propagation Block. 
Each block is initialized with two buffers, which will be filled with intermediate features. For $l$\ts{th} block, we denote them as current buffer $B^{0, l}  \in \mathbb{R}^{C_f \times H_f \times W_f}$ and past buffer $B^{-1, l}  \in \mathbb{R}^{f \times H_f \times W_f}$. We denote the relative temporal index as the superscript $0,-1$.

During the forward inference at time step $i\geq N$, a future feature $Z^{i-l, l}  \in \mathbb{R}^{C_f \times H_f \times W_f}$ is fed into $l$\ts{th} buffered block. The superscript index $i-l$ denotes latency of $l$ frames caused by $l$ blocks. More details about latency and the operation for $i < N$ at will be discussed later in Sec.\ref{section:Pipeline-style Inference Scheme}.
As shown in Figure~\ref{fig:Bidirectional Buffer Block}, The forward operation is composed of three steps:

\textbf{(a) Fusion of new input and buffered features.}
The temporal fusion can be represented as 
\begin{eqnarray}
Z^{i-l-1, l}_{fused} =& B^{-1, l} \oplus B^{0, l}_{[f:-f]} \oplus Z^{i-l, l}_{[-f:]},
\label{equ:temporal_fusion}
\end{eqnarray}
where two buffers are filled with features $B^{-1, l}=Z^{i-l-2, l}_{[0:f]}$, $B^{0, l}=Z^{i-l-1, l}$, which are cached in the last time step $i-1$. We will discuss the features in buffers in (c).

\textbf{(b) Feature extraction.} The feature extraction operation can be any general 2d operator, which is implemented as 2d convolution:
\begin{eqnarray}
Z^{i-l-1, l+1} =& Conv(Z^{i-l-1, l}_{fused}),
\end{eqnarray}
where $Z^{i-l-1, l+1}$ is the input for the next $(l+1)$\ts{th} Bidirectional Buffer Block.

\textbf{(c) Update the buffered feature in the memory.}
At the end of timestep $i$ for $l$\ts{th} block, we update the intermediate feature in the fixed buffer to reuse the previous computation. 
\begin{eqnarray}
        B^{-1,l} = & B^{0,l}_{[0:f]} \\
        B^{0,l} = & Z^{i-l, l}
\end{eqnarray}

After the above three steps, the temporal index of output ($i-l-1$ for $Z^{i-l-1,l+1}$) has moved backward by $1$ frame compared with the original input ($i-l$ for $Z^{i-l, l}$), since most of $Z^{i-l-1, l}_{fused}$'s channels are from $B^{0, l}=Z^{i-l-1, l}$.

\subsection{Pipeline-style Inference}
\label{section:Pipeline-style Inference Scheme}
\begin{table}[t]
\small
\centering
\renewcommand{\arraystretch}{1.2}
\caption{\small{Pipeline-Style inference at different time step.}\vspace{-5pt}}
\resizebox{\linewidth}{!}{
\begin{tabular}{@{}l@{\hspace{4mm}}l@{\hspace{4mm}}l@{\hspace{4mm}}c@{}}
\toprule
\textrm{Time step} & {Input} & {Buffered features} & {Output}\\

\midrule

{0} & {$x_0$}   &   {} & {None}\\ 
{1} & {$x_1$}   &  {$Z^{0,1}$} & {None}\\ 
{2} & {$x_2$}   &  {$Z^{1,1},Z^{0,2}$} & {None}\\ 
$\hdots$ \\
{$N$} & {$x_{N}$}  & {$Z^{N-1,1},Z^{N-2,2} \hdots, Z^{1,N-1}$} & {$Z^{0,N}=y_{0}$} \\ 
{$i \geq N$} &  {$x_{i}$} & {$Z^{i-1,1}, Z^{i-2,2}, \hdots, Z^{i-N+1,N-1}$}  & {$Z^{i-N,N}=y_{i-N}$} \\
\bottomrule
\vspace{-2em}
\end{tabular}
}
\label{table:Pipeline-style Inference Scheme}
\vspace{-1em}
\end{table}
 A single Bidirectional Buffer Block can be seen as a sliding window with size 3, which has 1 frame temporal latency. 
 We consider the first Bidirectional Buffer Block for ease of description.
 At time step $i=0$, both buffers $(B^{-1, 1}$ and $B^{0, 1})$ in 1\ts{st} block are empty, which means this block does not have enough information for temporal fusion. Thus, it caches the input $Z^{0,1}$ to $B^{0,1}$, and the whole pipeline exits. At time step $i=1$, together with the new input, the block is activated by information from two time steps. It fills $B^{-1,l}$ with zeros and conduct forward temporal fusion to extract feature $Z^{0,1}$ for $x_0$.
 Table~\ref{table:Pipeline-style Inference Scheme} shows the operation and intermediate feature at each time step.
 For time step $i<N$, each input frame activates one deeper Bidirectional Buffer Block $B^{0,i}$ in our network. 
Therefore, when the first denoised frame $y_0$ comes out, there are already $N+1$ frames fed into our inference block. At a general time step $i\geq N$, the network takes $x_i$ as input. $l$\ts{th} block takes $Z^{i-l, l}$ as input and produce $Z^{i-l-1, l+1}$.
The whole pipeline generates and caches $N$ features. 
Finally, the network produces a clean frame output $y_{i-N}$ at the end of the time step $i$.
For the last $N$ frames in the entire video sequence, we feed dummy zero tensors into the pipeline to get the clean output of the last few frames. Our supplement provides more details about the inference pipeline.

\subsection{Analysis}
\textbf{Running time.}
FastDVDnet~\cite{Tassano2020FastDVDNet} is a sliding-window-based framework with a window size of 5, which is composed of 2 U-Nets with the smallest channel 32. FastDVDnet fuses the three adjacent noisy frames at each U-Net's input layer. Thus, the computation cost for each output is $3+1=4$ times that of a single U-Net. In our work, BSVD-32 uses two similar U-Nets with the same channel setting as the convolution backbone. Instead of feeding a clip of frames, we feed in a single frame at any time step. Therefore, the computation cost of our method for each frame is $1+1=2$ times of U-Net. As shown in Table~\ref{table:davis_set8_baseline_comparison}, our BSVD-32 reduces more than $50\%$ of runtime compared with FastDVDnet.

\textbf{Memory.}
In bidirectional-RNN and MIMO frameworks~\cite{liang2022vrt,Xiang2022ReMoNet,chan2021basicvsr}, memory consumption is typically proportional to the clip length $T_{clip}$ at inference time, which impedes the processing of the long video on devices with limited resources. Thus, the whole video is divided into multiple short sequences to fit the memory size on the user device. 
However, for low-level denoising tasks, this will lead to a quality drop on the boundary frames in a video clip due to the loss of the marginal feature (Sec.~\ref{section:Streaming inference vs parallel inference}).
Unlike MIMO frameworks, our buffers work in a pipeline that consumes frames one by one. Therefore, our method can keep a constant run-time memory, which depends only on the number of Bidirectional Buffer Blocks $N$.

\textbf{Temporal receptive field.}
The temporal receptive field of a single Bidirectional Propagation Block is $3$.
During the inference of the neural network, the receptive field of temporal shift will be accumulated like 1-D convolution. 
In our framework, we apply $N=16$ layers of Bidirectional Propagation Block in the network.
Therefore, the accumulated temporal receptive field of our BSVD is 33, which is larger than previous window-based methods (e.g., 7 for PaCNet~\cite{Vaksman2021Patch} and 5 for FastDVDnet~\cite{Tassano2020FastDVDNet}) by a large margin.


\begin{figure*}[t]
\centering
\vspace{-10pt}
\includegraphics[width=0.99\linewidth]{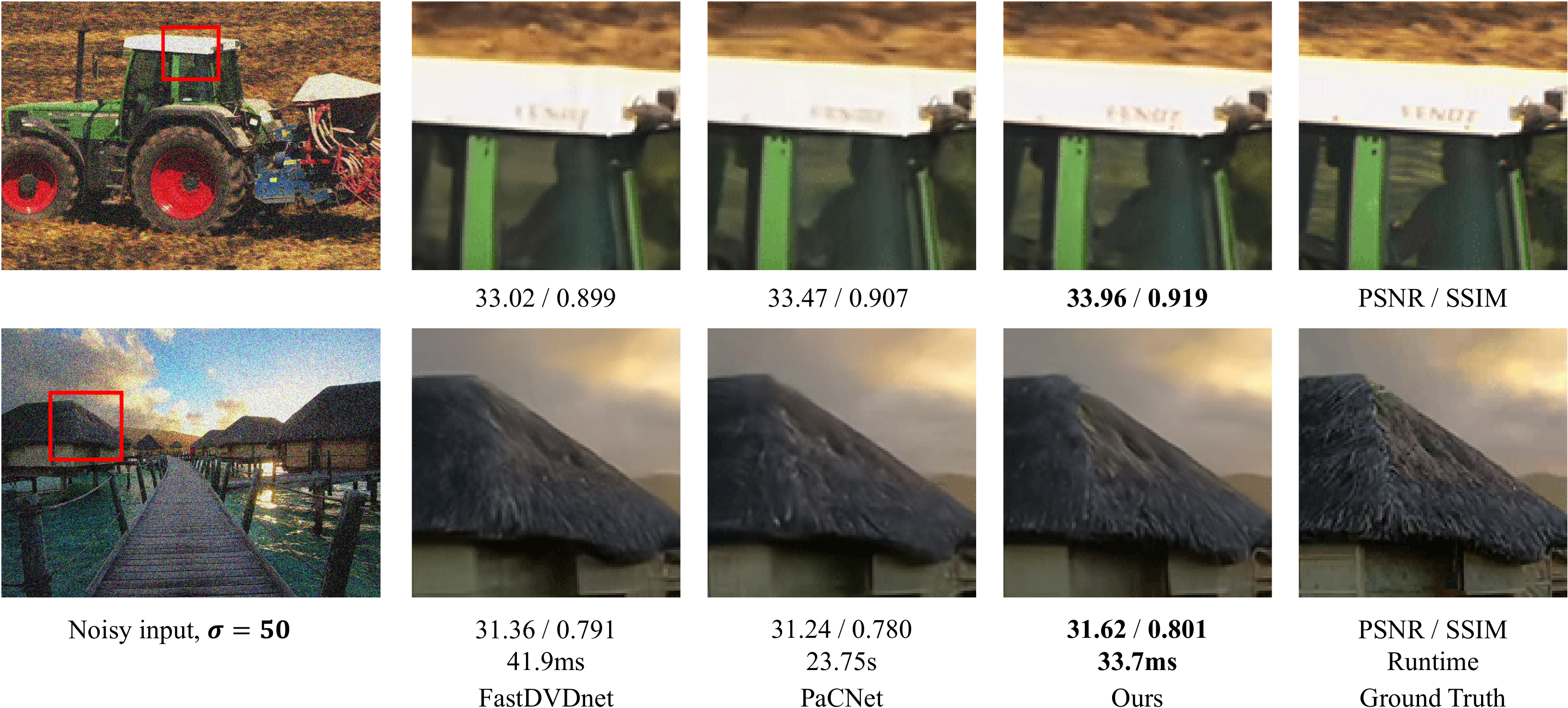}
\caption{Qualitative comparison on Set8 dataset. Ours reconstructs more high-frequency details in shorter running time.}
\label{fig:visual_set8}
\end{figure*}

\begin{table*}[t]
\centering
\caption{Quantitative comparisons of PSNR (dB) and runtime on the test set of DAVIS and Set8. C and G represent CPU and GPU time cost, respectively. $10, 20, 30, 40, 50$ represents the $\sigma$ of test data. 
We show the averaged inference time per frame with the resolution of 960 $\times$ 540 in PyTorch framework with FP16 precision. Our method outperforms previous methods on average PSNR and runtime. \vspace{-10pt}}
\renewcommand{\arraystretch}{0.9}
\resizebox{\textwidth}{!}{
\begin{tabular}{@{}l@{\hspace{3.5mm}}r@{\hspace{2mm}}*{12}{c@{\hspace{2mm}}}c@{}}

\toprule[1pt]
& & \multicolumn{6}{c}{DAVIS}& &\multicolumn{6}{c}{Set8} \\ 
\textrm{Model} & {Runtime(s)} & {$10$} &  {$20$} &  {$30$} &  {$40$} &   {$50$} &{Average}& & {$10$} &  {$20$} &  {$30$} &  {$40$} &   {$50$} &{Average}\\ 
\cmidrule[0.6pt](r{0.6em}){1-2}
\cmidrule[0.6pt] (r{0.2em}) {3-8}
\cmidrule[0.6pt]{10-15}

 {VNLB}~\cite{Arias2018VNLB}     & {420.0 (C)}     & {38.85} & {35.68}  & {33.73} & {32.32} & {31.13} & {34.34 } & & {\textbf{37.26}}  & {33.72}  & {31.74} & {30.39} & {29.24} &{32.47}\\ 
 {V-BM4D}~\cite{Maggioni2012VBM4D}   & {156.0 (C)}     & {37.58} & {33.88}  & {31.65} & {30.05} & {28.80} &{32.39} & & {36.05} & {32.19}  & {30.00} & {28.48} & {27.33} & {30.81} \\ 
 {VNLnet}~\cite{Davy2018Non-Local}   & {1.104 (G)}       & {35.83} & {34.49}  & {-} & {32.32} & {31.43} &{-}& & {37.10}  & {33.88}  & {-} & {30.55} & {29.47} & {-}\\ 
 {DVDnet}~\cite{tassano2019dvdnet}   &  {4.731 (C+G)}  & {38.13} & {35.70}  & {34.08} & {32.86} & {31.85}  & {34.52 }& & {36.08} & {33.49}  & {31.79} & {30.55} & {29.56} & {32.29}\\ 
 {FastDVDnet~\cite{Tassano2020FastDVDNet}} & {0.0419 (G)}    & {38.71}   & {35.77} & {34.04}  & {32.82} & {31.86} &{34.64} & & {36.44}         & {33.43} & {31.68}  & {30.46} & {29.53} & {32.31}\\ 
 {PaCNet~\cite{Vaksman2021Patch}} & {23.75 (G)}    & {\textbf{39.97}}   & {\textbf{36.82}} & {34.79}  & {33.34} & {32.20} &{35.42} & & {37.06} & {\textbf{33.94}} & {32.05}  & {30.70} & {29.66} & {32.68}\\ 
 \midrule
 {BSVD-32 (ours)}&  {0.0163 (G)} & {39.47} & {36.36} & {34.57} & {33.31}  & {32.34} & {35.21}& &{36.49}      & {33.56} & {31.85} & {30.65}  & {29.74} & {32.46}\\ 
 {BSVD-64 (ours)}&  {\textbf{0.0337} (G)} & {39.81} & {\textbf{36.82}} & {\textbf{35.09}} & {\textbf{33.86}}  & {\textbf{32.91}} & {\textbf{35.70 }}& &{36.74} & {33.83} & {\textbf{32.14}} & {\textbf{30.97}} & {\textbf{30.06}} & \textbf{32.75}\\ 
\bottomrule[1pt]

\end{tabular}
}
\label{table:davis_set8_baseline_comparison}
\end{table*}

\section{Experiments}

\subsection{Datasets and Settings}
To evaluate our method, we use both RGB images with synthetic noise~\cite{Khoreva2018Video} and raw images with real-world noise~\cite{yue2020supervised}. 

\textbf{DAVIS and Set8 datasets.}
For RGB images with synthetic noise, we follow the data preparation of FastDVDnet~\cite{Tassano2020FastDVDNet}. 
The clean patches
are randomly sampled from the training set of the DAVIS~\cite{Khoreva2018Video}. 
The noisy patches are generated by adding additional white Gaussian noise (AWGN) of $\sigma \in [5, 50]$ to clean patches.
For models trained on the DAVIS train set, we evaluate them on both Set8 and DAVIS test sets~\cite{Tassano2020FastDVDNet}.
We follow the FastDVDnet to limit all test video sequences to the first 85 frames.

\textbf{CRVD dataset}.
For real-world raw images, we use the dataset collected by~\citeauthor{yue2020supervised}~\cite{yue2020supervised}. It is composed of one synthetic dataset (SRVD) generated from MOT~\cite{Milan2016MOT16AB} and one captured dataset (CRVD). The real raw noise is often assumed to conform to a heteroskedastic Gaussian distribution with signal-dependent variance:
\begin{equation}
\label{eq: het_gaussian_noise_model}
\sigma^{2}(\mathbf{x})=a \mathbf{x}+b,
\end{equation}
where $a, b \in \mathbb{R}$ are parameters for shot and read noise respectively~\cite{Maggioni2021Efficient}.
5 possible pairs of $(a, b)$ map to 5 different ISO settings.
Following previous works~\cite{yue2020supervised, Maggioni2021Efficient}, we train our model with SRVD and CRVD scenes 1 to 6, and test on CRVD scenes 7 to 11. 

\vspace{-5pt}
\subsection{Baselines}
We compare our method with state-of-the-art methods, including raw data baselines~\cite{Maggioni2021Efficient,yue2020supervised,Tassano2020FastDVDNet}, and RGB image baselines~\cite{Maggioni2012VBM4D,Tassano2020FastDVDNet,tassano2019dvdnet,Arias2018VNLB,Vaksman2021Patch,Xiang2022ReMoNet,Sheth2021UDVD}.
For evaluation on RGB images with Gaussian noise, we train our model with the same L2 loss as FastDVDnet~\cite{Tassano2020FastDVDNet}, and compare it with the quantitative results of~\cite{Maggioni2012VBM4D,Tassano2020FastDVDNet,tassano2019dvdnet,Arias2018VNLB} in the FastDVDnet paper. 
For experiments on raw images with real-world noise, we use the evaluation metric from 
RViDeNet~\cite{yue2020supervised} and train with L1 loss, 
following EMVD~\cite{Maggioni2021Efficient}.
In Table~\ref{table:crvd_baseline_comparison}, we quantitatively compare with the results of all baselines from the EMVD~\cite{Maggioni2021Efficient} paper. Since the code of EMVD is not open-sourced, we test the efficiency of EMVD using an unofficial implementation\footnote{\url{https://github.com/Baymax-chen/EMVD}}.


\begin{figure*}[t]
\centering
\includegraphics[width=1.0\linewidth]{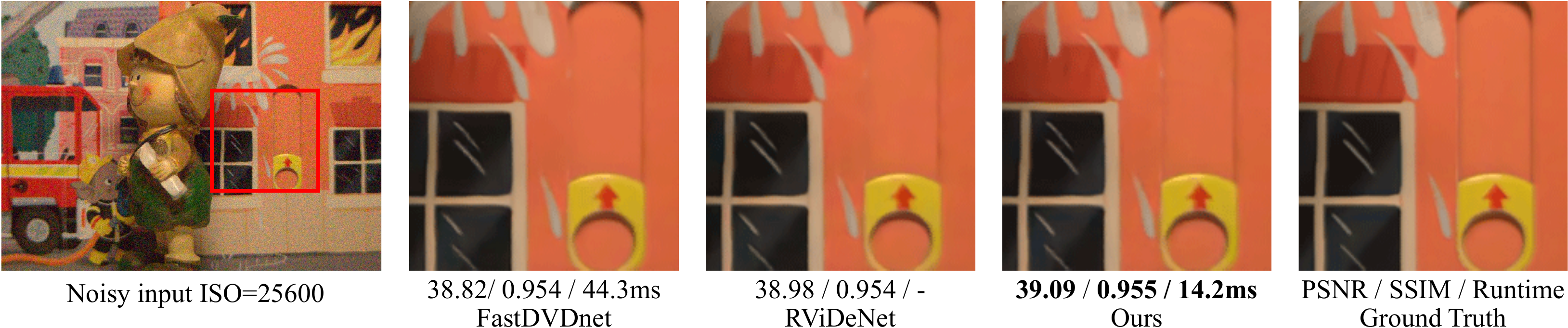}
\caption{Qualitative Comparison of real noisy frames from CRVD dataset. 
}
\label{fig:visual_crvd}
\end{figure*}

\begin{table*}[t]
\centering
\caption{Quantitative comparisons of PSNR for blind denoising. 
}
\renewcommand{\arraystretch}{0.9}
\resizebox{\textwidth}{!}{
\begin{tabular}{@{}l@{\hspace{3.5mm}}r@{\hspace{2mm}}*{12}{c@{\hspace{2mm}}}c@{}}

\toprule[1pt]
& & \multicolumn{6}{c}{DAVIS}& &\multicolumn{6}{c}{Set8} \\ 
\textrm{Model} & {Runtime(s)} & {$10$} &  {$20$} &  {$30$} &  {$40$} &   {$50$} &{Average}& & {$10$} &  {$20$} &  {$30$} &  {$40$} &   {$50$} &{Average}\\ 
\cmidrule[0.6pt](r{0.6em}){1-2}
\cmidrule[0.6pt] (r{0.2em}) {3-8}
\cmidrule[0.6pt]{10-15}

{ReMoNet~\cite{Xiang2022ReMoNet}} & {-} & 38.97 & 35.77 & 33.93 & 32.64 & 31.65 & 34.59 &  & 36.29 & 33.34 & 31.59 & 30.37 & 29.44 & 32.21\\ 
{UDVD~\cite{Sheth2021UDVD}} & {0.9561} & - & 34.99 & 33.86 & 32.61 & 31.63 & - & ~ & - & 33.25 & {31.86} & {30.62} & {29.63} & - \\ 
{BSVD-32-blind (ours)}&  \textbf{0.0162} & {39.31} & {36.23} & {34.46} & {33.21} & {32.25} & {35.09} & & 36.23 & {33.40} & 31.71 & 30.53 & 29.62 & 32.30\\ 
{BSVD-64-blind (ours)}&  {0.0335} & \textbf{39.68} & \textbf{36.66} & \textbf{34.91} & \textbf{33.68} & \textbf{32.72} & \textbf{35.53} & ~ & \textbf{36.54} & \textbf{33.70} & \textbf{32.02} & \textbf{30.85} & \textbf{29.95} & \textbf{32.61}\\ 
\bottomrule[1pt]

\end{tabular}
}
\label{table:davis_set8_baseline_comparison_blind_denoising}
\end{table*}

\subsection{Results}

\textbf{Non-blind denoising on RGB images. }
Table~\ref{table:davis_set8_baseline_comparison} shows the quantitative comparison of our method with baselines on the DAVIS and Set8 datasets. 
The inference time of VNLB, V-BM4D is obtained from the FastDVDnet paper. 
These CPU-based baselines are extremely slower than GPU-based methods. 
For fairness, we test the inference time of the most related baselines~\cite{Vaksman2021Patch,Tassano2020FastDVDNet,tassano2019dvdnet,Davy2018Non-Local} and our method in PyTorch framwork on the same NVIDIA RTX3090 GPU with FP16 precision.

The results in Table~\ref{table:davis_set8_baseline_comparison} indicates that our model surpasses all previous methods in terms of overall denoising performance with a much lower runtime. 
Compared with the best fidelity baseline PaCNet~\cite{Vaksman2021Patch}, our BSVD-64 achieve a 0.28dB improvement in average PSNR on the DAVIS dataset and 0.13dB on Set8 with $700\times$ speedup during inference. 
In addition, our GPU memory cost (3.09GB) is much lower than the cost of PaCNet (10.58GB).
Specifically, our method has significant advantages at $\sigma=50$ (e.g., 0.71dB improvement on DAVIS, 0.4dB on Set8). PaCNet exploits explicit search and fusion of nearest neighbor patches in the RGB pixel domain, which consumes an extremely long runtime and lacks matching of deep features. In contrast, our method implicitly conducts the alignment and fusion through convolution on multi-scale features, which proves to be more efficient with better fidelity. Figure~\ref{fig:visual_set8} also demonstrates that our method restores more high-frequency details on the characters and the roof.
Our BSVD-64 processes the video streams of resolution $960\times540$ at 30Hz framerate, which achieves real-time performance.

\begin{table}[t]
\small
\centering
\renewcommand{\arraystretch}{1.0}
\caption{Quantitative comparisons of PSNR and SSIM on the CRVD test set. 
We test the averaged computation cost per frame on RGGB raw data with resolution $960\times540$.
}
\resizebox{0.47\textwidth}{!}{
\begin{tabular}{@{}l@{\hspace{1mm}}c@{\hspace{1.5mm}}c@{\hspace{1.5mm}}c@{\hspace{1.5mm}}c@{}}
\toprule[1pt]
\textrm{Model}  & {Time(ms)} & {GFLOPs} & {raw} 
&  {sRGB}\\ 

\midrule
{RViDeNet~\cite{yue2020supervised}}  & {-}  &   {1965.0} & {44.08 / 0.9881} & {40.03 / 0.9802}\\ 
{FastDVDnet~\cite{Tassano2020FastDVDNet}}  & {44.3} &  {665.0} & {44.30} / 0.9891 & {39.91 / 0.9812}\\ 
{EMVD~\cite{Maggioni2021Efficient}}  & {24.0} & {79.5}  & {44.05 / 0.9890} & {39.53 / 0.9796}\\ 
\midrule
{BSVD-16 (ours)}  & {\textbf{9.9}} & {\textbf{78.76}} & {44.10 / 0.9884} & {40.17 / 0.9804}  \\ 
{BSVD-24 (ours)}& {14.2} & {175.46}  & {\textbf{44.39} / \textbf{0.9894}} & {\textbf{40.48} / \textbf{ 0.9820}}  \\
\bottomrule
\end{tabular}
}
\label{table:crvd_baseline_comparison}
\vspace{-10pt}
\end{table}

Furthermore, we half the channels in BSVD-64 to BSVD-32 as a lightweight version.
Compared with real-time SOTA FastDVDnet~\cite{Tassano2020FastDVDNet}, our BSVD-32 has more than $60\%$ reduction in runtime, with a 0.57dB increase of average PSNR on the DAVIS test set and 0.15dB on the Set8.





\textbf{Blind denoising on DAVIS and Set8 datasets.}
Our method is also applicable for blind video denoising without the noise map as input. We compare ours with previous blind denoising SOTA UDVD~\cite{Sheth2021UDVD}, and a concurrent work ~\cite{Xiang2022ReMoNet}. As shown in Table~\ref{table:davis_set8_baseline_comparison_blind_denoising}, our BSVD-64-blind shows better fidelity.


\textbf{Non-blind denoising on CRVD dataset.}
In Table~\ref{table:crvd_baseline_comparison}, we compare with baselines on real-world raw data.
Since the real-time baseline EMVD~\cite{Maggioni2021Efficient} has a very low computation cost, we shrink the channel of our model and train BSVD-24 and BSVD-16.
Compared with FastDvDnet, our BSVD-24 produces better images with $3\times$ speedup.
 Compared with EMVD, our BSVD-16 has 58.3\% shorter inference time, and achieves better image fidelity in both raw and sRGB domains.
Our model consists of only 2d convolutions, without special designs for raw data prior, such as attention, depth-wise separable convolution in EMVD. Thus, BSVD-16 has shorter runtime with similar GFLOPs as EMVD.
As shown in Figure~\ref{fig:visual_crvd}, our results have sharper edges on CRVD dataset. More results are shown in the supplement.
\subsection{Application of Buffer in FastDVDnet.}
Our buffer-based pipeline-style inference can be applied to the existing method.
FastDVDnet is a two-stage sliding-window-based method that conducts temporal fusion at the input layer of each U-Net.
We utilize the pre-trained checkpoint and buffer the intermediate feature during each forward inference,
which modifies the original computation graph into pipeline style.
As a result, we half the runtime from 42ms to 23ms with the same image fidelity as the original implementation. More details are in the supplement.







\section{Ablation Studies}

\begin{figure*}[t]
\centering
\begin{tabular}{@{}c@{\hspace{15mm}}c@{}}

\includegraphics[width=0.5\linewidth]{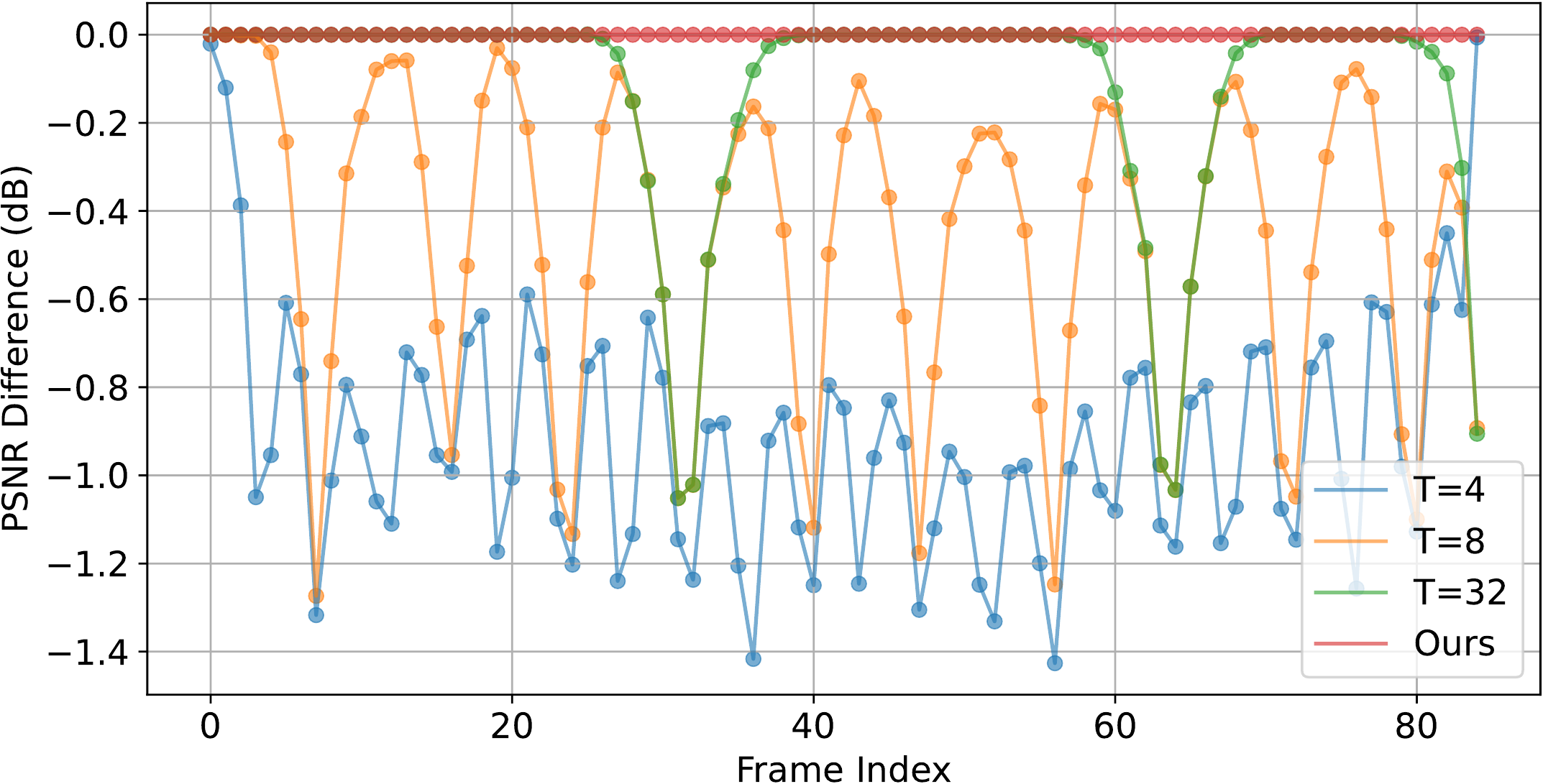}&
\includegraphics[width=0.36\linewidth]{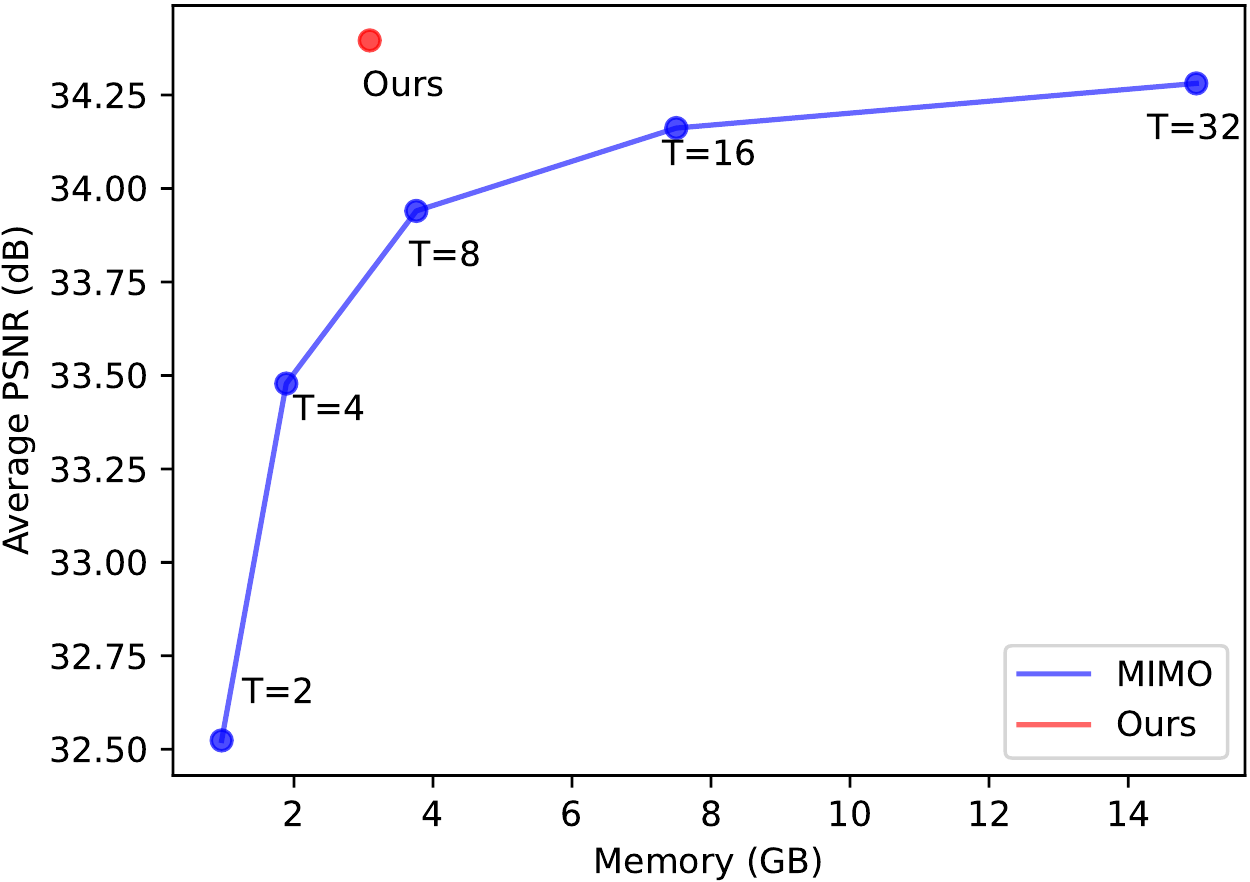} \\

(a) PSNR difference at each temporal index & (b) Average PSNR and memory consumption

\end{tabular}
\vspace{-6pt}
\caption{Fidelity and memory comparison between our pipeline framework and MIMO for a denoised video sequence $sunflower$ from Set8 at $\sigma = 50$. 
(a) Our inference method solves the PSNR drop on the two edges of the clips with up to 1dB improvement.
(b) Our framework consumes lower memory than $T_{clip}=8$ with a 0.45dB improvement at average PSNR.
}
\label{fig:buffer_ablation_quantitative}
\end{figure*}

\begin{figure}[ht]
\centering
\includegraphics[width=0.98\linewidth]{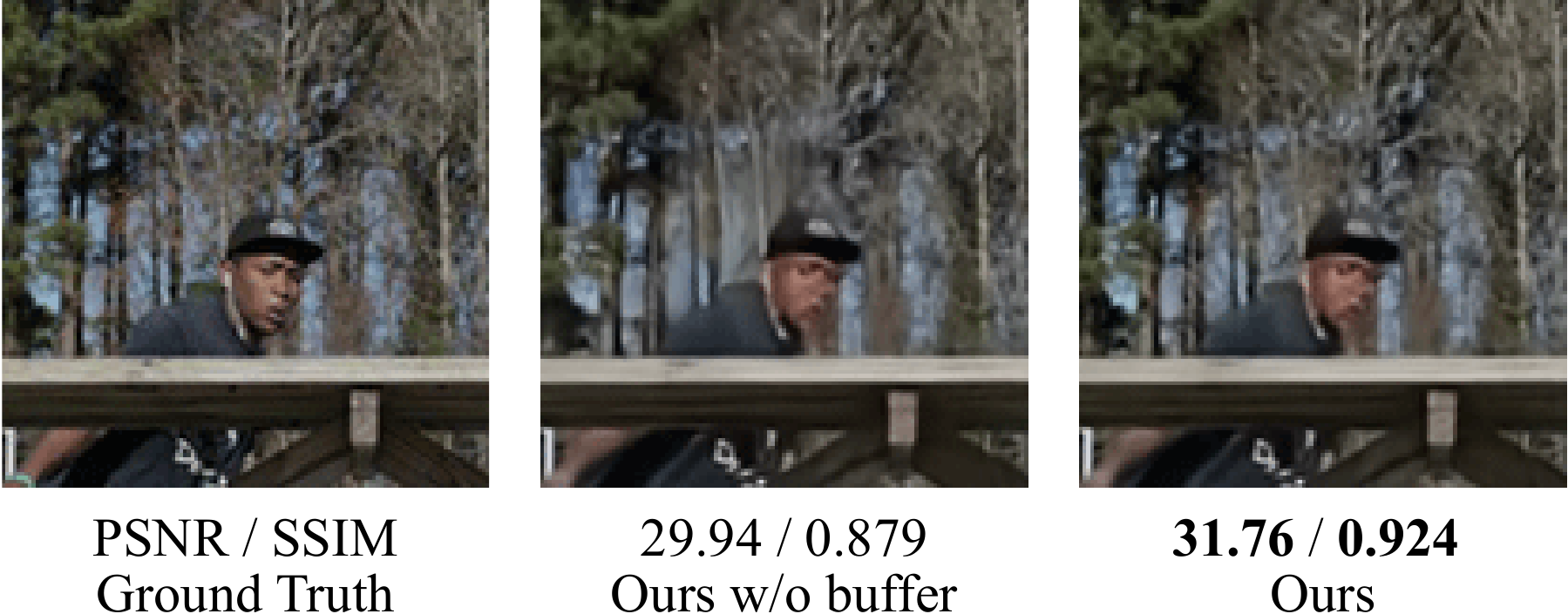}
\vspace{-6pt}
\caption{Qualitative ablation study of inference method.} 
\label{fig:buffer_ablation_qualitative}
\end{figure}

\begin{table}[t]
\centering
\renewcommand{\arraystretch}{1.2}
\caption{Quantitative ablation study based on our BSVD-64. The ``Ave'' denotes the average PSNR of $\sigma \in \{10,20,30,40,50\}$. Due to the space limit, we omit the results for $\sigma \in \{20,40\}$. \vspace{-5pt}}
\renewcommand{\arraystretch}{0.9}
\resizebox{0.47\textwidth}{!}{
\begin{tabular}{@{}l@{\hspace{1mm}}*{8}{c@{\hspace{1mm}}}r@{}}

\toprule[1pt]
& \multicolumn{4}{c}{DAVIS}& &\multicolumn{4}{c}{Set8} \\ 
\textrm{Model} & {$10$} &  {$30$} &  {$50$} &{Ave}& & {$10$} &  {$30$} & {$50$} &{Ave}\\ 
\cmidrule[0.6pt](r{0.6em}){1-1}
\cmidrule[0.6pt] (r{0.2em}) {2-5}
\cmidrule[0.6pt]{7-10}
 {Ours-Uni~\cite{Kondratyuk2021movinets}}& 39.40  & 34.40  & 32.14 & 35.06 & ~ & 36.51  & 31.70 & 29.54 & 32.34 \\


 
 {Ours-BN} & 39.14 & 34.57 & 32.33 & 35.13 & & 36.50 & 31.86 & 29.72 & 32.46 \\
  {Ours-U-Net~\cite{Ronneberger2015U-Net}}
& 39.54  & 34.66 & 32.44 & 35.30 & & 36.56 & 31.87 & 29.76 & 32.49 \\ 
 {Ours-MIMO~\cite{Lin2019TSM}} & 39.68 & 34.83 & 32.61 & 35.46 & & 36.68 & 31.98 & 29.86 & 32.60  \\
 \midrule
 {Ours} & \textbf{39.81} &  {\textbf{35.09}}   & {\textbf{32.91}} & {\textbf{35.70 }}& & \textbf{36.74}  & {\textbf{32.14}} & {\textbf{30.06}} & \textbf{32.75}\\ 
\bottomrule[1pt]
\end{tabular}
}
\label{table:davis_set8_baseline_comparison_ablation_study}
\vspace{-12pt}
\end{table}




\subsection{Pipeline Inference vs MIMO}
\label{section:Streaming inference vs parallel inference}
We modify our BSVD-64 into a MIMO framework using temporal shift operation~\cite{Lin2019TSM} with the same model parameters. We set the clip size $T_{clip}=8$, which is close to previous works (e.g., 5 for FastDVDnet~\cite{Tassano2020FastDVDNet} and ReMoNet~\cite{Xiang2022ReMoNet}, 7 for PaCNet~\cite{Vaksman2021Patch}). 

As shown in table~\ref{table:davis_set8_baseline_comparison_ablation_study}, our buffer-based pipeline inference framework brings 0.24dB improvement in average PSNR on DAVIS test set. 
To further analyze our image fidelity, we demonstrate the PSNR difference at each frame index
with clip size $T_{clip}=4,8,32$ on the video $sunflower$ from Set8 dataset in Figure~\ref{fig:buffer_ablation_quantitative}(a). The MIMO framework suffers significant fidelity degradation (up to 1 dB in PSNR) at the boundary of clips (e.g., 32\ts{nd} frame). Although increasing $T_{clip}$ improves the fidelity of the middle frames in the clip, it can not achieve obvious improvement on the two temporal edges of the clip.
As shown in Figure~\ref{fig:buffer_ablation_quantitative}(b), increasing the $T_{clip}$ of MIMO framework results in linear growth of memory consumption.
In contrast, our framework processes the video continuously in the constant memory. 
Figure~\ref{fig:buffer_ablation_qualitative} provides another qualitative comparison with $T_{clip}=8$ MIMO framework at 32th frame of the ``skate-jump'' sequence from DAVIS. Our bidirectional buffer improves the sharpness of the tree, especially in areas close to the moving player.

\subsection{Bidirectional Feature Propagation}
\label{section:Bidirectional_feature_propagation}
\textbf{Bidirectional vs unidirectional temporal fusion. }
 MoViNet \cite{Kondratyuk2021movinets} propose a buffer-based unidirectional temporal fusion for video steam recognition. 
 In Table~\ref{table:davis_set8_baseline_comparison_ablation_study}, we modify our bidirectional fusion to unidirectional temporal fusion with stream buffer~\cite{Kondratyuk2021movinets}:
\begin{eqnarray}
Z^{i, l+1}_{fused} =& Conv(B^{-1, l} \oplus Z^{i, l}_{[2f:]}).
\end{eqnarray}
Then, update the buffer $B^{-1,l}$ with $Z^{i,l}_{[0:2f]}$.
The unidirectional version of our method "Ours-Uni" does not need future frames and thus does not have frame latency. However, the results in table~\ref{table:davis_set8_baseline_comparison_ablation_study} show a large gap of 0.64dB between bidirectional and unidirectional fusion.
For low-level video tasks, the bidirectional fusion can be more advantageous since the future frames also have similar and helpful features to be utilized by each frame.


\textbf{Number of buffer blocks N.}
In Table~\ref{table:ablation of N}, we add buffer blocks at pixel resolution or remove them at downsampled resolution, using the same backbone.
Larger $N$ brings a larger receptive field (RF) at the cost of longer latency during online inference. 
The result shows that pixel-level fusion is not necessary. Thus, we only fuse the downsampled features, which can be more robust against noise.

\begin{table}[t]
\small
\centering
\caption{Ablation study for the number of buffer blocks $N$. \vspace{-5pt}
}
\renewcommand{\arraystretch}{0.9}
\resizebox{0.47\textwidth}{!}{
\begin{tabular}{@{}l@{\hspace{1mm}}*{3}c@{\hspace{0.5mm}} c@{\hspace{1mm}}c@{}}
\toprule[1pt]
\textrm{Model} & {$N$} & {RF} &Scale &  {DAVIS-Ave} &  {Set8-Ave}\\ 

\midrule
{Ours-w/o fusion} & {0} & 1 &  {-} & 34.01 & 31.63 \\
{Ours-only Pixel} & {2} & 5 & { pixel} & 34.03 & 31.66 \\
 {Ours-with Pixel} & {24} & 49 & {Down + pixel} & 35.12 & 32.35 \\
 \midrule
 {BSVD-64 (ours)} &  {16} & 33 & {Down}& \textbf{35.70} & \textbf{32.75} \\
\bottomrule
\vspace{-2em}
\end{tabular}
}
\label{table:ablation of N}
\end{table}

\subsection{Backbone}


\textbf{Batch normalization. }
We train a W-Net with BN layers and tune other hyperparameters. BN layers decreases the average PSNR by 0.57dB on the DAVIS test set, as shown in Table~\ref{table:davis_set8_baseline_comparison_ablation_study}.

\textbf{W-Net vs U-Net.}
We replace our backbone with a U-Net with increased channels, such that the new model has similar FLOPs and runtime as W-Net. As shown in Table~\ref{table:davis_set8_baseline_comparison_ablation_study}, W-Net outperforms U-Net by 0.40dB.
This result indicates that a second U-Net for refinement may help to remove the remaining noise.

\section{Conclusion}
We propose a SOTA streaming video denoising method BSVD that outperforms existing methods on videos with synthetic and real noise in both inference speed and image fidelity.
Our pipeline-style inference with Bidirectional Buffer Blocks allows bidirectional temporal fusion for online streaming video processing,
which is proved to be more effective than unidirectional fusion. In addition, we solve the degradation of clip edges, which exists in MIMO frameworks. Our method is effective for both non-blind and blind denoising, and is also general for similar architectures.
Extensive experiments on public datasets have demonstrated the effectiveness of our method.


\bibliographystyle{ACM-Reference-Format}
\bibliography{ref}


\begin{thebibliography}{35}


\ifx \showCODEN    \undefined \def \showCODEN     #1{\unskip}     \fi
\ifx \showDOI      \undefined \def \showDOI       #1{#1}\fi
\ifx \showISBNx    \undefined \def \showISBNx     #1{\unskip}     \fi
\ifx \showISBNxiii \undefined \def \showISBNxiii  #1{\unskip}     \fi
\ifx \showISSN     \undefined \def \showISSN      #1{\unskip}     \fi
\ifx \showLCCN     \undefined \def \showLCCN      #1{\unskip}     \fi
\ifx \shownote     \undefined \def \shownote      #1{#1}          \fi
\ifx \showarticletitle \undefined \def \showarticletitle #1{#1}   \fi
\ifx \showURL      \undefined \def \showURL       {\relax}        \fi
\providecommand\bibfield[2]{#2}
\providecommand\bibinfo[2]{#2}
\providecommand\natexlab[1]{#1}
\providecommand\showeprint[2][]{arXiv:#2}

\bibitem[Arias and Morel(2018)]%
        {Arias2018VNLB}
\bibfield{author}{\bibinfo{person}{Pablo Arias} {and}
  \bibinfo{person}{Jean{-}Michel Morel}.} \bibinfo{year}{2018}\natexlab{}.
\newblock \showarticletitle{Video Denoising via Empirical Bayesian Estimation
  of Space-Time Patches}.
\newblock \bibinfo{journal}{\emph{J. Math. Imaging Vis.}}
  (\bibinfo{year}{2018}).
\newblock


\bibitem[Buades et~al\mbox{.}(2011)]%
        {Buades2011Non-Local}
\bibfield{author}{\bibinfo{person}{Antoni Buades}, \bibinfo{person}{Bartomeu
  Coll}, {and} \bibinfo{person}{Jean{-}Michel Morel}.}
  \bibinfo{year}{2011}\natexlab{}.
\newblock \showarticletitle{Non-Local Means Denoising}.
\newblock \bibinfo{journal}{\emph{Image Process. Line}}  \bibinfo{volume}{1}
  (\bibinfo{year}{2011}).
\newblock


\bibitem[Chan et~al\mbox{.}(2021)]%
        {chan2021basicvsr}
\bibfield{author}{\bibinfo{person}{Kelvin~C.K. Chan}, \bibinfo{person}{Xintao
  Wang}, \bibinfo{person}{Ke Yu}, \bibinfo{person}{Chao Dong}, {and}
  \bibinfo{person}{Chen~Change Loy}.} \bibinfo{year}{2021}\natexlab{}.
\newblock \showarticletitle{BasicVSR: The Search for Essential Components in
  Video Super-Resolution and Beyond}. In \bibinfo{booktitle}{\emph{CVPR}}.
\newblock


\bibitem[Dabov et~al\mbox{.}(2007)]%
        {Dabov2007imagedenoising}
\bibfield{author}{\bibinfo{person}{Kostadin Dabov}, \bibinfo{person}{Alessandro
  Foi}, \bibinfo{person}{Vladimir Katkovnik}, {and} \bibinfo{person}{Karen
  Egiazarian}.} \bibinfo{year}{2007}\natexlab{}.
\newblock \showarticletitle{Image denoising by sparse 3D transform-domain
  collaborative filtering}.
\newblock \bibinfo{journal}{\emph{IEEE Transactions on Image Processing}}
  (\bibinfo{year}{2007}).
\newblock


\bibitem[Davy et~al\mbox{.}(2018)]%
        {Davy2018Non-Local}
\bibfield{author}{\bibinfo{person}{Axel Davy}, \bibinfo{person}{Thibaud Ehret},
  \bibinfo{person}{Jean~Michel Morel}, \bibinfo{person}{Pablo Arias}, {and}
  \bibinfo{person}{Gabriele Facciolo}.} \bibinfo{year}{2018}\natexlab{}.
\newblock \showarticletitle{{Non-Local Video Denoising by CNN}}.
\newblock \bibinfo{journal}{\emph{arXiv}}  \bibinfo{volume}{abs/1811.12758}
  (\bibinfo{year}{2018}).
\newblock


\bibitem[Khoreva et~al\mbox{.}(2018)]%
        {Khoreva2018Video}
\bibfield{author}{\bibinfo{person}{Anna Khoreva}, \bibinfo{person}{Anna
  Rohrbach}, {and} \bibinfo{person}{Brent Schiele}.}
  \bibinfo{year}{2018}\natexlab{}.
\newblock \showarticletitle{Video Object Segmentation with Referring
  Expressions}. In \bibinfo{booktitle}{\emph{ECCV}}.
\newblock


\bibitem[Kingma and Ba(2015)]%
        {Kingma2015Adam}
\bibfield{author}{\bibinfo{person}{Diederik~P. Kingma} {and}
  \bibinfo{person}{Jimmy Ba}.} \bibinfo{year}{2015}\natexlab{}.
\newblock \showarticletitle{Adam: {A} Method for Stochastic Optimization}. In
  \bibinfo{booktitle}{\emph{ICLR}}.
\newblock


\bibitem[Kondratyuk et~al\mbox{.}(2021)]%
        {Kondratyuk2021movinets}
\bibfield{author}{\bibinfo{person}{Dan Kondratyuk}, \bibinfo{person}{Liangzhe
  Yuan}, \bibinfo{person}{Yandong Li}, \bibinfo{person}{Li Zhang},
  \bibinfo{person}{Mingxing Tan}, \bibinfo{person}{Matthew Brown}, {and}
  \bibinfo{person}{Boqing Gong}.} \bibinfo{year}{2021}\natexlab{}.
\newblock \showarticletitle{MoViNets: Mobile Video Networks for Efficient Video
  Recognition}. In \bibinfo{booktitle}{\emph{CVPR}}.
\newblock


\bibitem[Lebrun et~al\mbox{.}(2013)]%
        {Lebrun2013Nonlocal}
\bibfield{author}{\bibinfo{person}{Marc Lebrun}, \bibinfo{person}{Antoni
  Buades}, {and} \bibinfo{person}{Jean{-}Michel Morel}.}
  \bibinfo{year}{2013}\natexlab{}.
\newblock \showarticletitle{A Nonlocal Bayesian Image Denoising Algorithm}.
\newblock \bibinfo{journal}{\emph{{SIAM} J. Imaging Sci.}}
  (\bibinfo{year}{2013}).
\newblock


\bibitem[Liang et~al\mbox{.}(2022)]%
        {liang2022vrt}
\bibfield{author}{\bibinfo{person}{Jingyun Liang}, \bibinfo{person}{Jiezhang
  Cao}, \bibinfo{person}{Yuchen Fan}, \bibinfo{person}{Kai Zhang},
  \bibinfo{person}{Rakesh Ranjan}, \bibinfo{person}{Yawei Li},
  \bibinfo{person}{Radu Timofte}, {and} \bibinfo{person}{Luc Van~Gool}.}
  \bibinfo{year}{2022}\natexlab{}.
\newblock \showarticletitle{VRT: A Video Restoration Transformer}.
\newblock \bibinfo{journal}{\emph{arXiv:2201.12288}} (\bibinfo{year}{2022}).
\newblock


\bibitem[Liang et~al\mbox{.}(2021)]%
        {liang2021swinir}
\bibfield{author}{\bibinfo{person}{Jingyun Liang}, \bibinfo{person}{Jiezhang
  Cao}, \bibinfo{person}{Guolei Sun}, \bibinfo{person}{Kai Zhang},
  \bibinfo{person}{Luc Van~Gool}, {and} \bibinfo{person}{Radu Timofte}.}
  \bibinfo{year}{2021}\natexlab{}.
\newblock \showarticletitle{SwinIR: Image Restoration Using Swin Transformer}.
  In \bibinfo{booktitle}{\emph{ICCV Workshops}}.
\newblock


\bibitem[Lim et~al\mbox{.}(2017)]%
        {Lim2017EDSR}
\bibfield{author}{\bibinfo{person}{Bee Lim}, \bibinfo{person}{Sanghyun Son},
  \bibinfo{person}{Heewon Kim}, \bibinfo{person}{Seungjun Nah}, {and}
  \bibinfo{person}{Kyoung~Mu Lee}.} \bibinfo{year}{2017}\natexlab{}.
\newblock \showarticletitle{Enhanced Deep Residual Networks for Single Image
  Super-Resolution}. In \bibinfo{booktitle}{\emph{CVPR Workshops}}.
\newblock


\bibitem[Lin et~al\mbox{.}(2019)]%
        {Lin2019TSM}
\bibfield{author}{\bibinfo{person}{Ji Lin}, \bibinfo{person}{Chuang Gan}, {and}
  \bibinfo{person}{Song Han}.} \bibinfo{year}{2019}\natexlab{}.
\newblock \showarticletitle{{TSM:} Temporal Shift Module for Efficient Video
  Understanding}. In \bibinfo{booktitle}{\emph{ICCV}}.
\newblock


\bibitem[Maggioni et~al\mbox{.}(2012)]%
        {Maggioni2012VBM4D}
\bibfield{author}{\bibinfo{person}{Matteo Maggioni}, \bibinfo{person}{Giacomo
  Boracchi}, \bibinfo{person}{Alessandro Foi}, {and} \bibinfo{person}{Karen~O.
  Egiazarian}.} \bibinfo{year}{2012}\natexlab{}.
\newblock \showarticletitle{Video Denoising, Deblocking, and Enhancement
  Through Separable 4-D Nonlocal Spatiotemporal Transforms}.
\newblock \bibinfo{journal}{\emph{IEEE Transactions on Image Processing}}
  (\bibinfo{year}{2012}).
\newblock


\bibitem[Maggioni et~al\mbox{.}(2021)]%
        {Maggioni2021Efficient}
\bibfield{author}{\bibinfo{person}{Matteo Maggioni}, \bibinfo{person}{Yibin
  Huang}, \bibinfo{person}{Cheng Li}, \bibinfo{person}{Shuai Xiao},
  \bibinfo{person}{Zhongqian Fu}, {and} \bibinfo{person}{Fenglong Song}.}
  \bibinfo{year}{2021}\natexlab{}.
\newblock \showarticletitle{Efficient Multi-Stage Video Denoising With
  Recurrent Spatio-Temporal Fusion}. In \bibinfo{booktitle}{\emph{CVPR}}.
\newblock


\bibitem[Maggioni et~al\mbox{.}(2013)]%
        {Maggioni2013NonlocalTF}
\bibfield{author}{\bibinfo{person}{Matteo Maggioni}, \bibinfo{person}{V.
  Katkovnik}, \bibinfo{person}{K. Egiazarian}, {and} \bibinfo{person}{A. Foi}.}
  \bibinfo{year}{2013}\natexlab{}.
\newblock \showarticletitle{Nonlocal Transform-Domain Filter for Volumetric
  Data Denoising and Reconstruction}.
\newblock \bibinfo{journal}{\emph{IEEE Transactions on Image Processing}}
  (\bibinfo{year}{2013}).
\newblock


\bibitem[Milan et~al\mbox{.}(2016)]%
        {Milan2016MOT16AB}
\bibfield{author}{\bibinfo{person}{Anton Milan}, \bibinfo{person}{L.
  Leal-Taix{\'e}}, \bibinfo{person}{I. Reid}, \bibinfo{person}{S. Roth}, {and}
  \bibinfo{person}{K. Schindler}.} \bibinfo{year}{2016}\natexlab{}.
\newblock \showarticletitle{MOT16: A Benchmark for Multi-Object Tracking}.
\newblock \bibinfo{journal}{\emph{ArXiv}}  \bibinfo{volume}{abs/1603.00831}
  (\bibinfo{year}{2016}).
\newblock


\bibitem[Mildenhall et~al\mbox{.}(2018)]%
        {Mildenhall2018Burst}
\bibfield{author}{\bibinfo{person}{Ben Mildenhall},
  \bibinfo{person}{Jonathan~T. Barron}, \bibinfo{person}{Jiawen Chen},
  \bibinfo{person}{Dillon Sharlet}, \bibinfo{person}{Ren Ng}, {and}
  \bibinfo{person}{Robert Carroll}.} \bibinfo{year}{2018}\natexlab{}.
\newblock \showarticletitle{Burst Denoising With Kernel Prediction Networks}.
  In \bibinfo{booktitle}{\emph{CVPR}}.
\newblock


\bibitem[Nah et~al\mbox{.}(2017)]%
        {SeungjunNah2017DeepMC}
\bibfield{author}{\bibinfo{person}{Seungjun Nah}, \bibinfo{person}{Tae~Hyun
  Kim}, {and} \bibinfo{person}{Kyoung~Mu Lee}.}
  \bibinfo{year}{2017}\natexlab{}.
\newblock \showarticletitle{Deep Multi-scale Convolutional Neural Network for
  Dynamic Scene Deblurring}. In \bibinfo{booktitle}{\emph{CVPR}}.
\newblock


\bibitem[Ronneberger et~al\mbox{.}(2015)]%
        {Ronneberger2015U-Net}
\bibfield{author}{\bibinfo{person}{Olaf Ronneberger}, \bibinfo{person}{Philipp
  Fischer}, {and} \bibinfo{person}{Thomas Brox}.}
  \bibinfo{year}{2015}\natexlab{}.
\newblock \showarticletitle{U-Net: Convolutional Networks for Biomedical Image
  Segmentation}. In \bibinfo{booktitle}{\emph{MICCAI}}.
\newblock


\bibitem[Sheth et~al\mbox{.}(2021)]%
        {Sheth2021UDVD}
\bibfield{author}{\bibinfo{person}{Dev~Yashpal Sheth}, \bibinfo{person}{Sreyas
  Mohan}, \bibinfo{person}{Joshua~L. Vincent}, \bibinfo{person}{Ramon
  Manzorro}, \bibinfo{person}{Peter~A. Crozier}, \bibinfo{person}{Mitesh~M.
  Khapra}, \bibinfo{person}{Eero~P. Simoncelli}, {and} \bibinfo{person}{Carlos
  Fernandez-Granda}.} \bibinfo{year}{2021}\natexlab{}.
\newblock \showarticletitle{Unsupervised Deep Video Denoising}. In
  \bibinfo{booktitle}{\emph{ICCV}}.
\newblock


\bibitem[Tassano et~al\mbox{.}(2019)]%
        {tassano2019dvdnet}
\bibfield{author}{\bibinfo{person}{Matias Tassano}, \bibinfo{person}{Julie
  Delon}, {and} \bibinfo{person}{Thomas Veit}.}
  \bibinfo{year}{2019}\natexlab{}.
\newblock \showarticletitle{Dvdnet: A fast network for deep video denoising}.
  In \bibinfo{booktitle}{\emph{ICIP}}.
\newblock


\bibitem[Tassano et~al\mbox{.}(2020)]%
        {Tassano2020FastDVDNet}
\bibfield{author}{\bibinfo{person}{Matias Tassano}, \bibinfo{person}{Julie
  Delon}, {and} \bibinfo{person}{Thomas Veit}.}
  \bibinfo{year}{2020}\natexlab{}.
\newblock \showarticletitle{{FastDVDNet: Towards real-time deep video denoising
  without flow estimation}}. In \bibinfo{booktitle}{\emph{CVPR}}.
\newblock


\bibitem[Vaksman et~al\mbox{.}(2021)]%
        {Vaksman2021Patch}
\bibfield{author}{\bibinfo{person}{Gregory Vaksman}, \bibinfo{person}{Michael
  Elad}, {and} \bibinfo{person}{Peyman Milanfar}.}
  \bibinfo{year}{2021}\natexlab{}.
\newblock \showarticletitle{Patch Craft: Video Denoising by Deep Modeling and
  Patch Matching}. In \bibinfo{booktitle}{\emph{ICCV}}.
\newblock


\bibitem[Wang et~al\mbox{.}(2019)]%
        {Wang2019EDVR}
\bibfield{author}{\bibinfo{person}{Xintao Wang}, \bibinfo{person}{Kelvin C.~K.
  Chan}, \bibinfo{person}{Ke Yu}, \bibinfo{person}{Chao Dong}, {and}
  \bibinfo{person}{Chen~Change Loy}.} \bibinfo{year}{2019}\natexlab{}.
\newblock \showarticletitle{{EDVR:} Video Restoration With Enhanced Deformable
  Convolutional Networks}. In \bibinfo{booktitle}{\emph{CVPR Workshops}}.
\newblock


\bibitem[Wang et~al\mbox{.}(2018)]%
        {esrgan}
\bibfield{author}{\bibinfo{person}{Xintao Wang}, \bibinfo{person}{Ke Yu},
  \bibinfo{person}{Shixiang Wu}, \bibinfo{person}{Jinjin Gu},
  \bibinfo{person}{Yihao Liu}, \bibinfo{person}{Chao Dong}, \bibinfo{person}{Yu
  Qiao}, {and} \bibinfo{person}{Chen~Change Loy}.}
  \bibinfo{year}{2018}\natexlab{}.
\newblock \showarticletitle{ESRGAN: Enhanced super-resolution generative
  adversarial networks}. In \bibinfo{booktitle}{\emph{ECCV Workshops}}.
\newblock


\bibitem[Wen et~al\mbox{.}(2017)]%
        {Wen2017Joint}
\bibfield{author}{\bibinfo{person}{Bihan Wen}, \bibinfo{person}{Yanjun Li},
  \bibinfo{person}{Luke Pfister}, {and} \bibinfo{person}{Yoram Bresler}.}
  \bibinfo{year}{2017}\natexlab{}.
\newblock \showarticletitle{Joint Adaptive Sparsity and Low-Rankness on the
  Fly: An Online Tensor Reconstruction Scheme for Video Denoising}. In
  \bibinfo{booktitle}{\emph{ICCV}}.
\newblock


\bibitem[Wen et~al\mbox{.}(2019)]%
        {Wen2019VIDOSAT}
\bibfield{author}{\bibinfo{person}{Bihan Wen}, \bibinfo{person}{Saiprasad
  Ravishankar}, {and} \bibinfo{person}{Yoram Bresler}.}
  \bibinfo{year}{2019}\natexlab{}.
\newblock \showarticletitle{{VIDOSAT:} High-Dimensional Sparsifying Transform
  Learning for Online Video Denoising}.
\newblock \bibinfo{journal}{\emph{IEEE Transactions on Image Processing}}
  (\bibinfo{year}{2019}).
\newblock


\bibitem[Xia and Kulis(2017)]%
        {Xia2017WNetAD}
\bibfield{author}{\bibinfo{person}{Xide Xia} {and} \bibinfo{person}{B. Kulis}.}
  \bibinfo{year}{2017}\natexlab{}.
\newblock \showarticletitle{W-Net: A Deep Model for Fully Unsupervised Image
  Segmentation}.
\newblock \bibinfo{journal}{\emph{ArXiv}}  \bibinfo{volume}{abs/1711.08506}
  (\bibinfo{year}{2017}).
\newblock


\bibitem[Xia et~al\mbox{.}(2020)]%
        {Xia2020Basis}
\bibfield{author}{\bibinfo{person}{Zhihao Xia}, \bibinfo{person}{Federico
  Perazzi}, \bibinfo{person}{Micha{\"e}l Gharbi}, \bibinfo{person}{Kalyan
  Sunkavalli}, {and} \bibinfo{person}{Ayan Chakrabarti}.}
  \bibinfo{year}{2020}\natexlab{}.
\newblock \showarticletitle{Basis prediction networks for effective burst
  denoising with large kernels}. In \bibinfo{booktitle}{\emph{CVPR}}.
\newblock


\bibitem[Xiang et~al\mbox{.}(2022)]%
        {Xiang2022ReMoNet}
\bibfield{author}{\bibinfo{person}{Liuyu Xiang}, \bibinfo{person}{Jundong
  Zhou}, \bibinfo{person}{Jirui Liu}, \bibinfo{person}{Zerun Wang},
  \bibinfo{person}{Haidong Huang}, \bibinfo{person}{Jie Hu},
  \bibinfo{person}{Jungong Han}, \bibinfo{person}{Yuchen Guo}, {and}
  \bibinfo{person}{Guiguang Ding}.} \bibinfo{year}{2022}\natexlab{}.
\newblock \showarticletitle{ReMoNet: Recurrent Multi-output Network for
  Efficient Video Denoising}. In \bibinfo{booktitle}{\emph{AAAI}}.
\newblock


\bibitem[Xu et~al\mbox{.}(2021)]%
        {Lu2021Efficient}
\bibfield{author}{\bibinfo{person}{Lu Xu}, \bibinfo{person}{Jiawei Zhang},
  \bibinfo{person}{Xuanye Cheng}, \bibinfo{person}{Feng Zhang},
  \bibinfo{person}{Xing Wei}, {and} \bibinfo{person}{Jimmy S.~J. Ren}.}
  \bibinfo{year}{2021}\natexlab{}.
\newblock \showarticletitle{Efficient Deep Image Denoising via Class Specific
  Convolution}. In \bibinfo{booktitle}{\emph{AAAI}}.
\newblock


\bibitem[Yue et~al\mbox{.}(2020)]%
        {yue2020supervised}
\bibfield{author}{\bibinfo{person}{Huanjing Yue}, \bibinfo{person}{Cong Cao},
  \bibinfo{person}{Lei Liao}, \bibinfo{person}{Ronghe Chu}, {and}
  \bibinfo{person}{Jingyu Yang}.} \bibinfo{year}{2020}\natexlab{}.
\newblock \showarticletitle{Supervised Raw Video Denoising with a Benchmark
  Dataset on Dynamic Scenes}. In \bibinfo{booktitle}{\emph{CVPR}}.
\newblock


\bibitem[Zamir et~al\mbox{.}(2022)]%
        {Zamir2021Restormer}
\bibfield{author}{\bibinfo{person}{Syed~Waqas Zamir}, \bibinfo{person}{Aditya
  Arora}, \bibinfo{person}{Salman Khan}, \bibinfo{person}{Munawar Hayat},
  \bibinfo{person}{Fahad~Shahbaz Khan}, {and} \bibinfo{person}{Ming-Hsuan
  Yang}.} \bibinfo{year}{2022}\natexlab{}.
\newblock \showarticletitle{Restormer: Efficient Transformer for
  High-Resolution Image Restoration}. In \bibinfo{booktitle}{\emph{CVPR}}.
\newblock


\bibitem[Zhang et~al\mbox{.}(2017)]%
        {zhang2017beyond}
\bibfield{author}{\bibinfo{person}{Kai Zhang}, \bibinfo{person}{Wangmeng Zuo},
  \bibinfo{person}{Yunjin Chen}, \bibinfo{person}{Deyu Meng}, {and}
  \bibinfo{person}{Lei Zhang}.} \bibinfo{year}{2017}\natexlab{}.
\newblock \showarticletitle{Beyond a gaussian denoiser: Residual learning of
  deep cnn for image denoising}.
\newblock \bibinfo{journal}{\emph{IEEE Transactions on Image Processing}}
  (\bibinfo{year}{2017}).
\newblock


\end{thebibliography}

\appendix
\newpage
\section{Implementation Details}

\subsection{Backbone Details}
Table~\ref{table:U-Net} shows details of a single U-Net in our BSVD-64 model.
TSM denotes Temporal Shift Module during training, while BBB stands for Bidirectional Buffer Block during inference.
“3×3, 64” denotes a 2d convolution with kernel size
3 and output channel 64. “ReLU6” denotes ReLU6 activation, whose output is clipped to be between $0$ to $6$.
Building blocks are shown in brackets, with the numbers of blocks stacked. Downsampling is performed using a convolution of stride 2 at the beginning of the downsampling block. 
After the last convolution in the upsampling block, we upsample the decoder feature using Pixel Shuffling. Then, skip-connections with intermediate features are conducted after upsampling. We stack two U-Nets as a W-Net for our backbone architecture. The output channel of the first U-Net and the input channel of the second U-Net are 64. During the training stage, we insert 16 Temporal Shift Module in our W-Net. During the inference stage, we replace each TSM with a Bidirectional Buffer Block.

\begin{table}[h]
\caption{\textbf{Architecture of our U-Net backbone in BSVD-64.}
}
\small
\centering
\renewcommand{\arraystretch}{1.0}
\resizebox{\linewidth}{!}{
\begin{tabular}{l@{\hspace{5mm}} l@{\hspace{5mm}} r}
\toprule
Stage & Building Block & Output Size\\

\midrule
\text{Input Layer} &
{$\begin{array}{l}
{\left[\begin{array}{l}
3 \times 3, 64 \\
\text{ReLU6} \\
\end{array}\right] \times 2 
} \\
\end{array}
$ }  
& ${H} \times {W} \times 64 $ \\

\hline
\text{Downsampling Block1}
& 
{
$\begin{array}{l}
    3 \times 3, 128, \text{stride=2} \\
    \text{ReLU6} \\
    \left[\begin{array}{l}
        \text{TSM } / \text{ BBB} \\
        3 \times 3, 128 \\
        \text{ReLU6} \\
    \end{array}\right] 
    \times 2 
\end{array}$
}
& $\frac{1}{2}H \times \frac{1}{2}W \times 128$  \\ 

\hline
\text{Downsampling Block2}
& 
{
$\begin{array}{l}
    3 \times 3, 256, \text{stride=2} \\
    \text{ReLU6} \\
    \left[\begin{array}{l}
        \text{TSM } / \text{ BBB} \\
        3 \times 3, 256 \\
        \text{ReLU6} \\
    \end{array}\right] 
    \times 2 
\end{array}$
}
& $\frac{1}{4}H \times \frac{1}{4}W \times 256$  \\

\hline
\text{Upsampling Block1}
& 
{
$\begin{array}{l}
    {\left[\begin{array}{l}
    \text{TSM } / \text{ BBB} \\
    3 \times 3, 256 \\ 
    \text{ReLU6} \\
    \end{array}\right] \times 2}  \\
    3 \times 3, 512 \\
    \text{PixelShuffle 2x} \\
\end{array}$
}
& $\frac{1}{2}H \times \frac{1}{2}W \times 128$  \\ 

\hline
\text{Upsampling Block2}
& 
{
$\begin{array}{l}
    {\left[\begin{array}{l}
    \text{TSM } / \text{ BBB} \\
    3 \times 3, 128 \\ 
    \text{ReLU6} \\
    \end{array}\right] \times 2}  \\
    3 \times 3, 256 \\
    \text{PixelShuffle 2x}
\end{array}$
}
& $H \times W \times 64$ \\
\hline
\small{Output Layer} &
$\begin{array}{l}
    3\times3, 64 \\
    \text{ReLU6} \\
    3\times3, 64 \text{ or } 3\times3, 3 \\
\end{array}$
& {$H \times W \times 64$ or $H \times W \times 3$} \\
\bottomrule
\end{tabular}
}
\label{table:U-Net}
\end{table}

\begin{figure*}[ht]
\centering
\includegraphics[width=\linewidth]{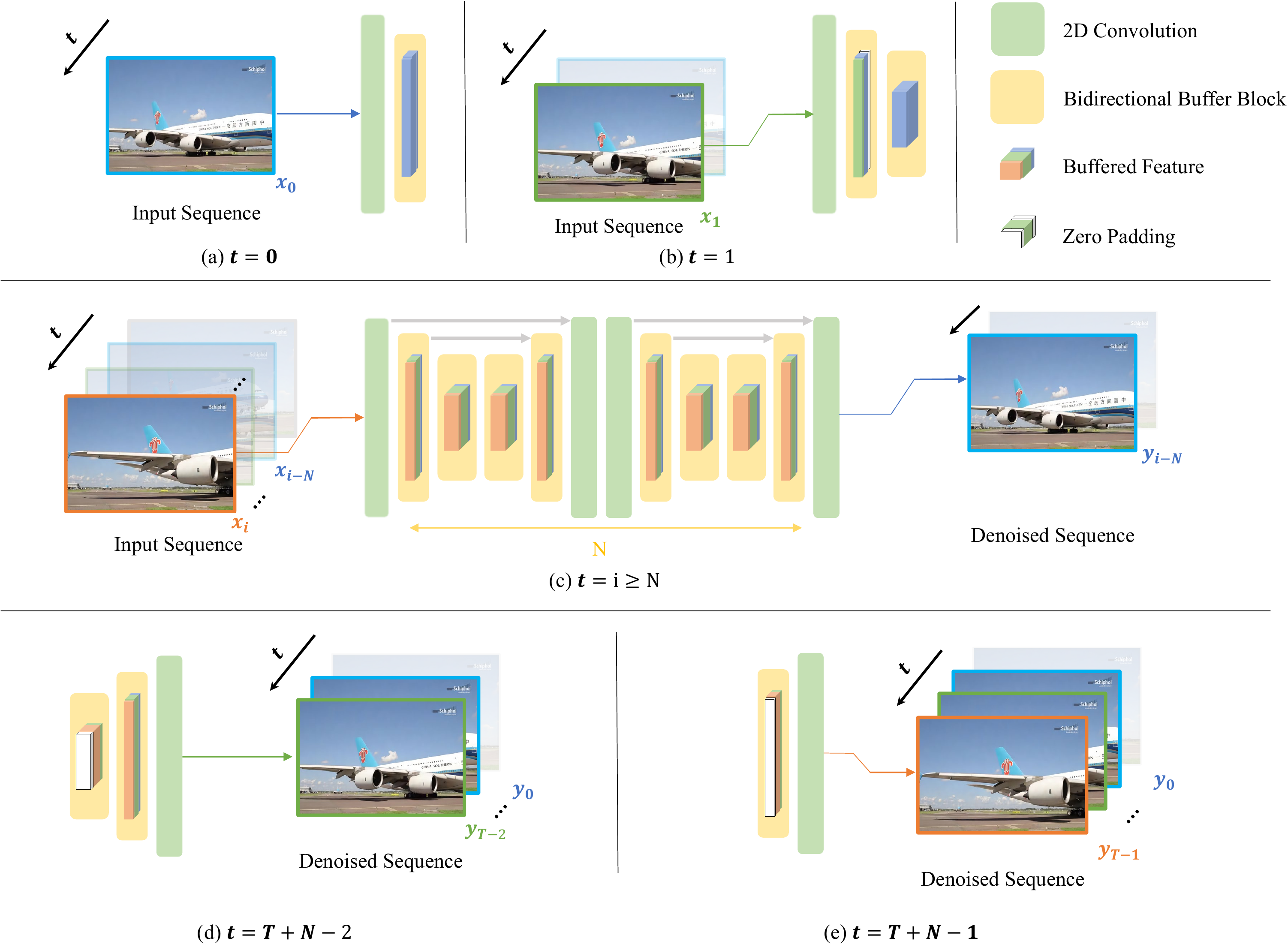}
\caption{ \textbf{Implementation of start and end in our pipeline}. For a long stream with $T$ frames, we use zero paddings at temporal index $t<N$ and $T\leq t \leq T+N-1$}
\label{fig:overview_start_ending}
\end{figure*}

\subsection{Edges of the Stream}
Fig.~\ref{fig:overview_start_ending} gives a visualization of our implementation at two edges of a long $T$-frame stream. We denote the temporal fusion equation as:
\begin{eqnarray}
Z^{i-l-1, l}_{fused} =& B^{-1, l} \oplus B^{0, l}_{[f:-f]} \oplus Z^{i-l, l}_{[-f:]},
\label{equ:temporal_fusion_supp}
\end{eqnarray}
We fill past buffer $B^{-1, l}$ with zero paddings at the temporal index $1\leq i<N$ (Fig.~\ref{fig:overview_start_ending}(b)). At the temporal index $T\leq i\leq T+N-1$, we feed dummy zero tensor $Z^{i-l, l}$ into the pipeline for the last N frames. (Fig.~\ref{fig:overview_start_ending}(d,e)).

\subsection{Implementation of Skip-Connection}
 Each skip-connection is conducted between two Bidirectional Buffer Blocks at different layers. Thus, there is a temporal latency between input and output during pipeline-style inference. We implement skip-connection as a first-in-first-out queue. The memory cost is proportional to the number of skipped layers, which is $O(N)$.

\subsection{Implementation of Buffer in FastDVDnet.}
We provide a supplementary description for the Sec. 4.4 in the main paper.
FastDVDnet~\cite{Tassano2020FastDVDNet} conducts temporal fusion at the input layer of each U-Net:
\begin{eqnarray}
Z^0_{fused} =&  Z ^ {-1} \oplus Z ^ {0} \oplus Z^ {+1},
\end{eqnarray}
which is similar to the temporal shift operation Equation. (1) in the main paper.
We modify the original computation graph into pipeline style
using a similar representation in Equation. (2-5) of the main paper:
\begin{eqnarray}
Z^{i-l-1,l}_{fused} &=& B^{-1,l} \oplus B^{0,l} \oplus Z^{i-l,l} \\
Z^{i-l-1, l+1} &=& \text{U-Net}(Z^{i-l-1,l}_{fused}) \\
B^{-1,l} &=& B^{0,l} \\
B^{0,l} &=& Z^{i-l,l},
\end{eqnarray}
where $l \in \{1,2\}$. As summarized in the Fig. 2 of the main paper, FastDVDNet is a sliding-window method without clip-edge-drop problem. However, its time complexity is $O(T_{clip}N)$, which is higher than $O(N)$ in our pipeline-style inference.
Using a pipeline-syle inference framework, we half the runtime from 42ms to 23ms with the same image fidelity as the original implementation.
\subsection{Training Details}
The model is trained with clips of batch size 16, temporal length $T_{clip}=11$ and spatial patch size $96 \times 96$ in each iteration. For each clip, We sample a random noise level $\sigma$ from the uniform distribution $U(5, 55)$ and augment the data with flipping and rotation. The training objective is optimized for $700,000$ iterations with Adam optimizer~\cite{Kingma2015Adam} of initial learning rate $1e-3$. The learning rate is decayed by a factor of $0.7$ for every $50,000$ iterations.

\section{More results}

\begin{figure}[ht]
\centering
\includegraphics[width=1.0\linewidth]{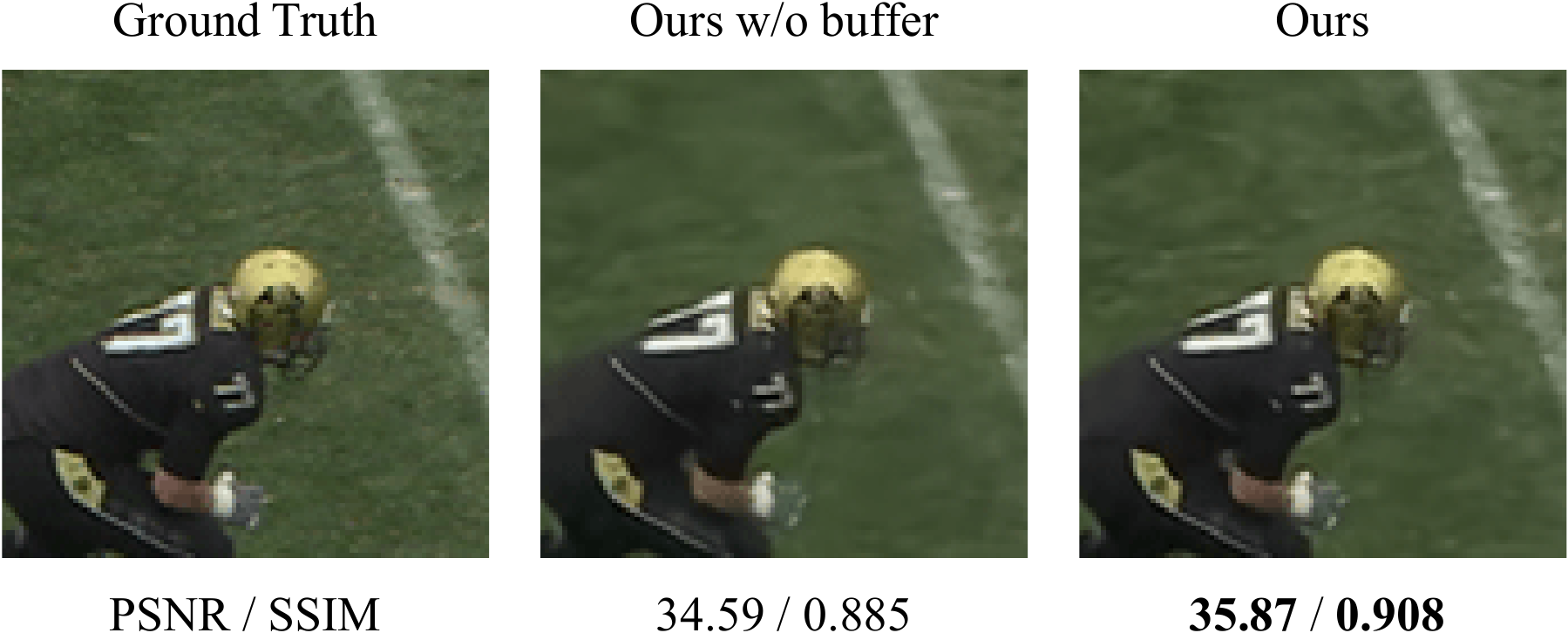}
\caption{Qualitative ablation of inference method}
\label{fig:ablation_pipeline2}
\end{figure}

\begin{figure}[ht]
\centering
\includegraphics[width=0.8\linewidth]{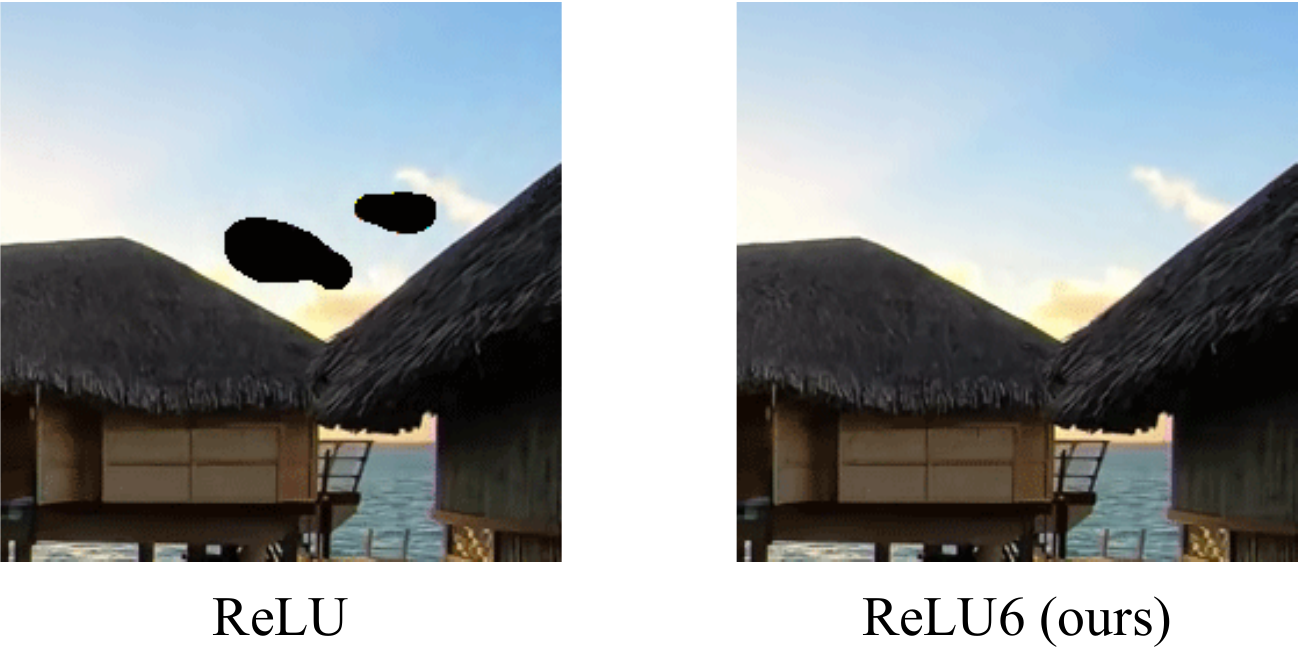}
\caption{{ReLU6 alleviates the artifacts caused by FP16 quantization.}}
\label{fig:quantization artifact}
\end{figure}

\begin{table*}[ht]
\centering
\caption{Quantitative ablation study of framework}
\renewcommand{\arraystretch}{1.0}
\resizebox{0.9\textwidth}{!}{
\begin{tabular}{@{}l@{\hspace{3.5mm}}*{12}{c@{\hspace{2mm}}}c@{}}

\toprule[1pt]
 & \multicolumn{6}{c}{DAVIS}& &\multicolumn{6}{c}{Set8} \\ 
\textrm{Model} 
& {$10$} &  {$20$} &  {$30$} &  {$40$} &   {$50$} &{Average}& & {$10$} &  {$20$} &  {$30$} &  {$40$} &   {$50$} &{Average}\\ 
\cmidrule[0.6pt](r{0.6em}){1-1}
\cmidrule[0.6pt] (r{0.2em}) {2-7}
\cmidrule[0.6pt]{9-14}
 {Ours-Unidirectional~\cite{Kondratyuk2021movinets}} & 39.40 & 36.22 & 34.40 & 33.12 & 32.14 & 35.06 & ~ & 36.51 & 33.47 & 31.70 & 30.48 & 29.54 & 32.34 \\
 {Ours-BN} 
 & 39.14 & 36.33 & 34.57 & 33.31 & 32.33 & 35.13 & & 36.50 & 33.57 & 31.86 & 30.65 & 29.72 & 32.46 \\
 {Ours-U-Net~\cite{Ronneberger2015U-Net}} 
& 39.54 & 36.45 & 34.66 & 33.41 & 32.44 & 35.30 & & 36.56 & 33.59 & 31.87 & 30.68 & 29.76 & 32.49 \\ 

 {Ours-MIMO~\cite{Lin2019TSM}} 
 & 39.68 & 36.61 & 34.83 & 33.58 & 32.61 & 35.46 & & 36.68 & 33.71 & 31.98 & 30.78 & 29.86 & 32.60  \\ 

 \midrule
 {BSVD-64(Ours)}
 & \textbf{39.81} & \textbf{36.82} & {\textbf{35.09}} & {\textbf{33.86}}  & {\textbf{32.91}} & {\textbf{35.70 }}& & \textbf{36.74} & \textbf{33.83} & {\textbf{32.14}} & {\textbf{30.97}} & {\textbf{30.06}} & \textbf{32.75}\\ 
\bottomrule[1pt]

\end{tabular}
}
\label{table:ablation_full}
\end{table*}

\begin{table*}[h]
\centering
\caption{Quantitative ablation study of shifted ratio $r$. $ r=8$ is the best setting among $\{4,6,8,16\}$}
\renewcommand{\arraystretch}{1.0}
\resizebox{0.9\textwidth}{!}{
\begin{tabular}{@{}l@{\hspace{3.5mm}}*{12}{c@{\hspace{2mm}}}c@{}}

\toprule[1pt]
 & \multicolumn{6}{c}{DAVIS}& &\multicolumn{6}{c}{Set8} \\ 
\textrm{Shifted ratio $r$} 
& {$10$} &  {$20$} &  {$30$} &  {$40$} &   {$50$} &{Average}& & {$10$} &  {$20$} &  {$30$} &  {$40$} &   {$50$} &{Average}\\ 
\cmidrule[0.6pt](r{0.6em}){1-1}
\cmidrule[0.6pt] (r{0.2em}) {2-7}
\cmidrule[0.6pt]{9-14}

4 & 39.74  & 36.72  & 34.97  & 33.74  & 32.78  & 35.59  & ~ & 36.73  & 33.79  & 32.09  & 30.90  & 30.00  & 32.70  \\ 
6 & 39.65  & 36.58  & 34.81  & 33.56  & 32.59  & 35.44  & ~ & 36.66  & 33.74  & 32.02  & 30.83  & 29.91  & 32.63  \\ 
8 & \textbf{39.81}  & \textbf{36.82}  & \textbf{35.09}  & \textbf{33.86}  & \textbf{32.91}  & \textbf{35.70}  & ~ & \textbf{36.74}  & \textbf{33.83}  & \textbf{32.14}  & \textbf{30.97}  & \textbf{30.06}  & \textbf{32.75}  \\ 
16 & 39.71  & 36.66  & 34.90  & 33.66  & 32.70  & 35.52  & ~ & 36.67  & 33.73  & 32.03  & 30.84  & 29.93  & 32.64 \\ 
\bottomrule[1pt]

\end{tabular}
}
\label{table:ablation_ratio}
\end{table*}

\begin{figure*}[ht]
\centering
\includegraphics[width=1.0\linewidth]{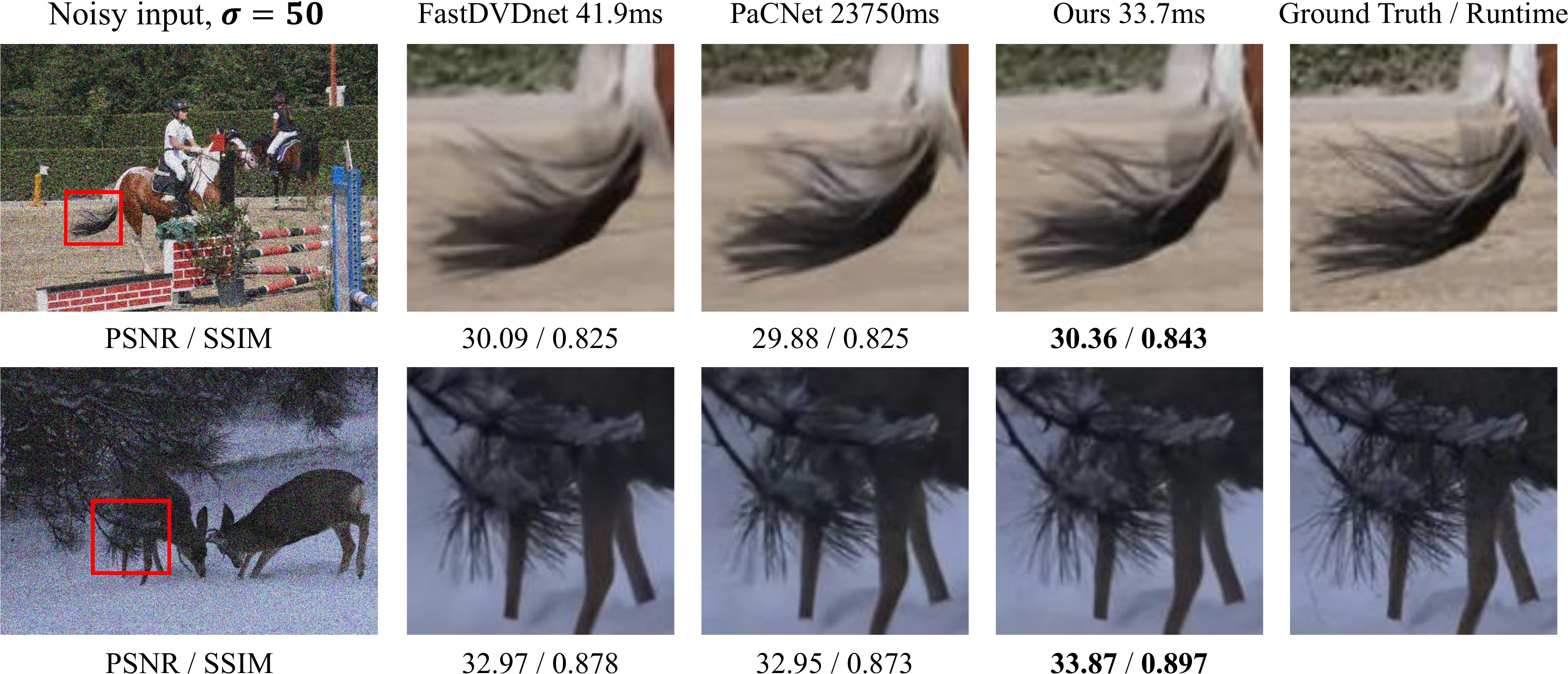}
\caption{Qualitative comparison on DAVIS dataset.}
\label{fig:visual_davis_supp}
\end{figure*}

\begin{figure*}[ht]
\centering
\includegraphics[width=1.0\linewidth]{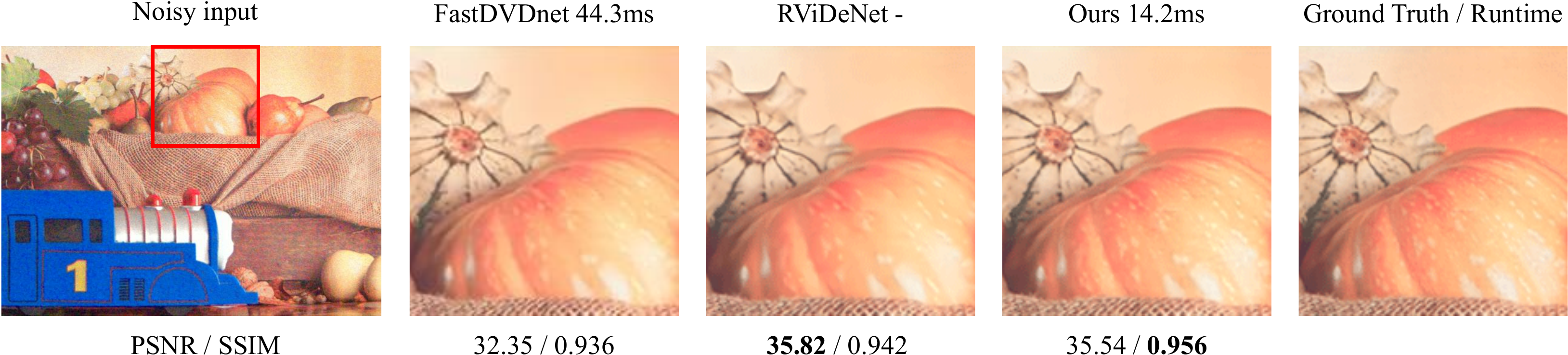}
\caption{Qualitative comparison on CRVD dataset}
\label{fig:visual_crvd_supp}
\end{figure*}

Fig.~\ref{fig:visual_davis_supp} and Fig.~\ref{fig:visual_crvd_supp} provide more qualitative comparison on DAVIS and CRVD dataset, respectively. Our method reconstructs more realistic high-frequency details (e.g., horsetail, snow on pine branches, spots on pumpkins). The source code and results of EMVD~\cite{Maggioni2021Efficient} and the concurrent work ReMoNet~\cite{Xiang2022ReMoNet} are not available. We only do quantitative comparisons with them. Since the source code of RViDeNet~\cite{yue2020supervised} is not compatible with CUDA version of our RTX 3090 GPU, we do not compare running time with RViDeNet.
There is another supplementary video for visualization of our algorithm and more results.


\section{More ablation Study}

Table~\ref{table:ablation_full} provides a supplement of all noise levels for the quantitative ablation study (Table 5) in the main paper. Bidirectional temporal fusion substantially improves the reconstructed quality. Moreover, a W-Net without batch normalization contributes to the performance. In addition, the designed pipeline-style inference also demonstrates better image quality than original MIMO framework.

For a TSM or Bidirectional Buffer Block with input feature channel $C_f$, we shift $f = \lfloor C_f/r\rfloor$ channels along the temporal dimension. We study the ratio of shifted channels in Table.~\ref{table:ablation_ratio}. We retrain our BSVD-64 with $r \in\{4,6,8,16\}$. The model of $r=8$ achieves the best result. Thus, we set $r=8$ in other experiments.
Fig.~\ref{fig:quantization artifact} shows an example of artifacts during the FP16 inference of a model with ReLU activation. To alleviate quantization artifacts, we replace the ReLU with ReLU6, which limits the maximum of activation to 6.

\begin{figure*}[h]
\centering
\includegraphics[width=0.98\linewidth]{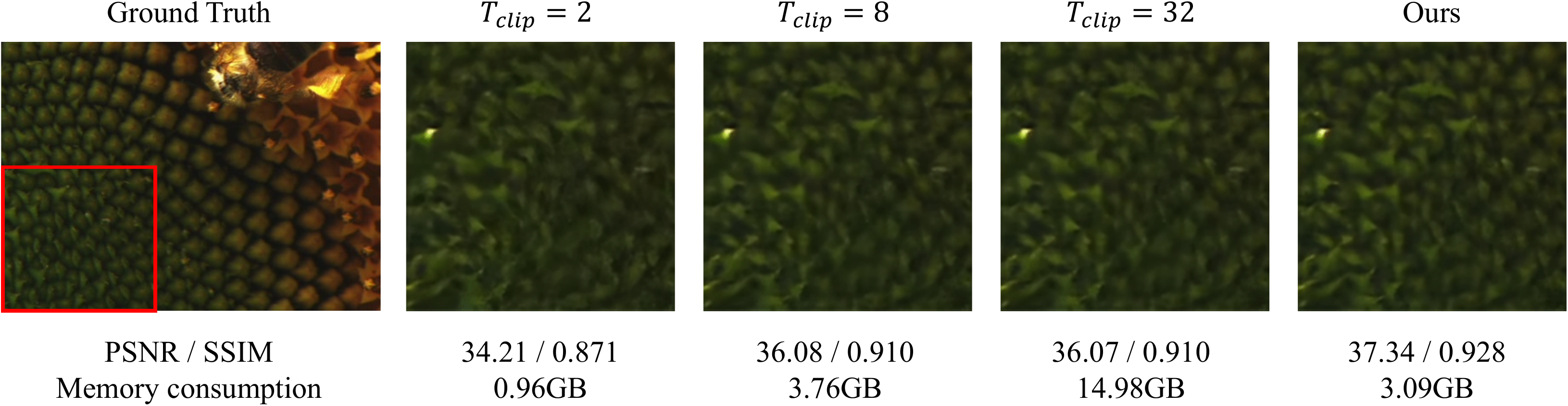}
\caption{Comparison with MIMO of different $T_{clip}$.}
\label{fig:T_ablation}
\end{figure*}

In Fig.~\ref{fig:T_ablation}, we compare our inference method with MIMO of different $T_{clip}$. We show the result on the clip boundaries. A larger $T_{clip}$ in MIMO framework increases the memory linearly and barely improves edge fidelity.

\end{document}